\documentclass{article}

\usepackage{amsmath,amsfonts,bm}

\def\Figref#1{Figure~\ref{#1}}

\def\Secref#1{Section~\ref{#1}}

\def\eqref#1{equation~\ref{#1}}

\def\1{\bm{1}}

\DeclareMathAlphabet{\mathsfit}{\encodingdefault}{\sfdefault}{m}{sl}
\SetMathAlphabet{\mathsfit}{bold}{\encodingdefault}{\sfdefault}{bx}{n}

\usepackage[sort,authoryear,round]{natbib}
\usepackage{fullpage}
\usepackage[utf8]{inputenc} 
\usepackage[T1]{fontenc}    
\usepackage{xcolor}
\usepackage{hyperref}       
\definecolor{myblue}{rgb}{0,0.2,0.8}
\hypersetup{ %
pdftitle={},
pdfauthor={},
pdfsubject={},
pdfkeywords={},
pdfborder=0 0 0,
pdfpagemode=UseNone,
colorlinks=true,
linkcolor=myblue,
citecolor=myblue,
filecolor=myblue,
urlcolor=myblue,
pdfview=FitH}
\usepackage{authblk}
\usepackage{url}            
\usepackage{booktabs}       
\usepackage{amsfonts}       
\usepackage{amsthm}
\usepackage{nicefrac}       
\usepackage{microtype}      

\usepackage{graphicx}
\usepackage{xspace}
\usepackage[varqu,varl,var0,scaled=0.97]{inconsolata}
\usepackage{caption}
\usepackage{overpic}
\usepackage{wrapfig}
\usepackage{tcolorbox}
\usepackage{enumitem}
\usepackage{threeparttable}
\usepackage{subfig}

\usepackage{listings}
\usepackage{color}
\usepackage{lipsum}

\usepackage{multirow}
\usepackage{makecell}
\usepackage{float}
\usepackage{csquotes}

\definecolor{dkgreen}{rgb}{0,0.6,0}
\definecolor{gray}{rgb}{0.5,0.5,0.5}
\definecolor{mauve}{rgb}{0.58,0,0.82}

\newcommand{\eg}{\emph{e.g.},\xspace}
\newcommand{\ie}{\emph{i.e.},\xspace}

\usepackage[capitalize,noabbrev]{cleveref}

\title{\bf{Breaking the Illusion: Real-world Challenges\\for Adversarial Patches in Object Detection}}
\author[1]{Jakob Schack$^*$}
\author[1]{Katarina Petrovic$^*$}
\author[1,2]{Olga Saukh}
\affil[1]{Graz University of Technology, Austria}
\affil[2]{Complexity Science Hub Vienna, Austria}
\affil[ ]{}
\affil[ ]{\texttt{jakob.schack@student.tugraz.at, katarina.petrovic@tugraz.at, saukh@tugraz.at}}

\begin{document}
\date{}
\maketitle

\def\thefootnote{*}\footnotetext{Both authors contributed equally to this research.}

\begin{abstract}
Adversarial attacks pose a significant threat to the robustness and reliability of machine learning systems, particularly in computer vision applications. This study investigates the performance of adversarial patches for the YOLO object detection network in the physical world. Two attacks were tested: a patch designed to be placed anywhere within the scene -- global patch, and another patch intended to partially overlap with specific object targeted for removal from detection -- local patch. Various factors such as patch size, position, rotation, brightness, and hue were analyzed to understand their impact on the effectiveness of the adversarial patches. The results reveal a notable dependency on these parameters, highlighting the challenges in maintaining attack efficacy in real-world conditions. Learning to align digitally applied transformation parameters with those measured in the real world still results in up to a 64\% discrepancy in patch performance. These findings underscore the importance of understanding environmental influences on adversarial attacks, which can inform the development of more robust defenses for practical machine learning applications.
\end{abstract}

\section{Introduction}

The rapid advancement of machine learning algorithms, particularly in computer vision domain, has revolutionized various applications, from autonomous driving to medical imaging and secure face recognition. However, the widespread adoption of these algorithms has brought to light significant security concerns, notably the susceptibility to adversarial attacks~\citep{szegedy2014,Costa_2024}. These attacks involve deliberate modifications to input data to mislead machine learning models into producing incorrect outputs.

A prominent example of a computer vision system is the YOLO (You Only Look Once)~\citep{Terven_2023} detection network, known for its high speed and accuracy. Despite its advantages, YOLO, like many other machine learning models, is vulnerable to adversarial attacks~\citep{choi2022adversarial}. Highly effective in digital environments, these can also degrade the performance of detection networks in real-world scenarios.

In addition to adversarial attacks, the robustness of machine learning models, including YOLO, is also challenged by environmental conditions such as weather, lighting, camera location, and viewpoint changes~\citep{ding2024sdniayolo}. Variations in lighting conditions can lead to overexposure or underexposure, which in turn can affect the model's ability to correctly detect and classify objects. Similarly, changes in the camera's location and viewpoint can introduce new perspectives and angles that the model may not have been trained on, further reducing its detection accuracy. Environmental conditions also affect the performance of adversarial patches~\citep{hartnett2022empirical}.

\begin{figure}[t]
    \centering
    \subfloat[Physically changed hue
    \label{fig:physical_clean}]{\includegraphics[width=0.35\linewidth]{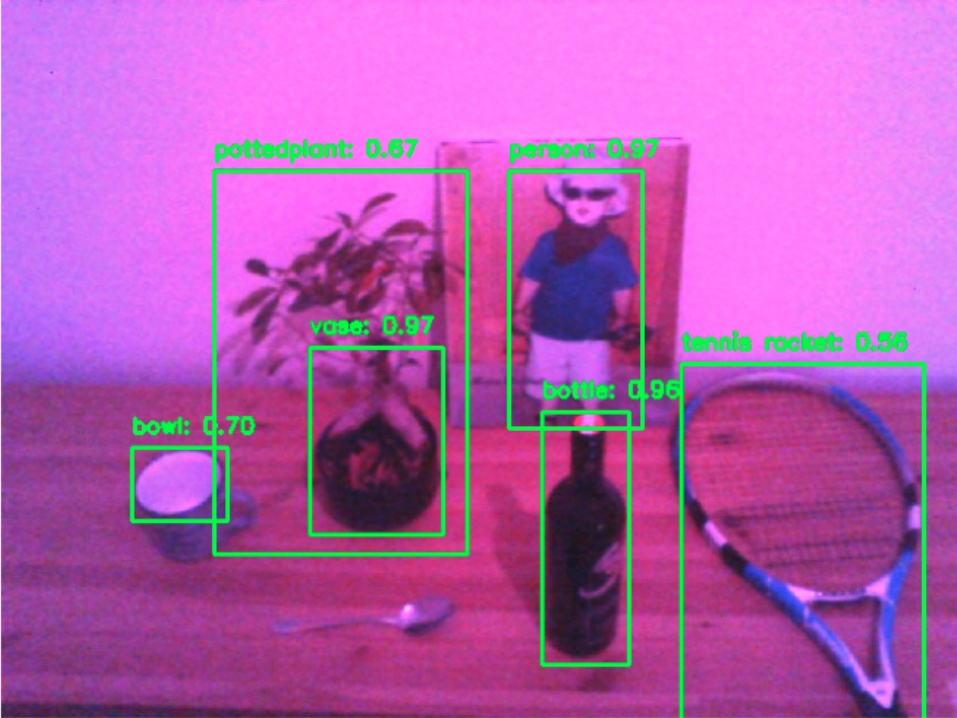}}
    \hspace*{10pt}
    \subfloat[As in (a) + patch = \color{red}{not effective}
    \label{fig:physical_patched}]{\includegraphics[width=0.35\linewidth]{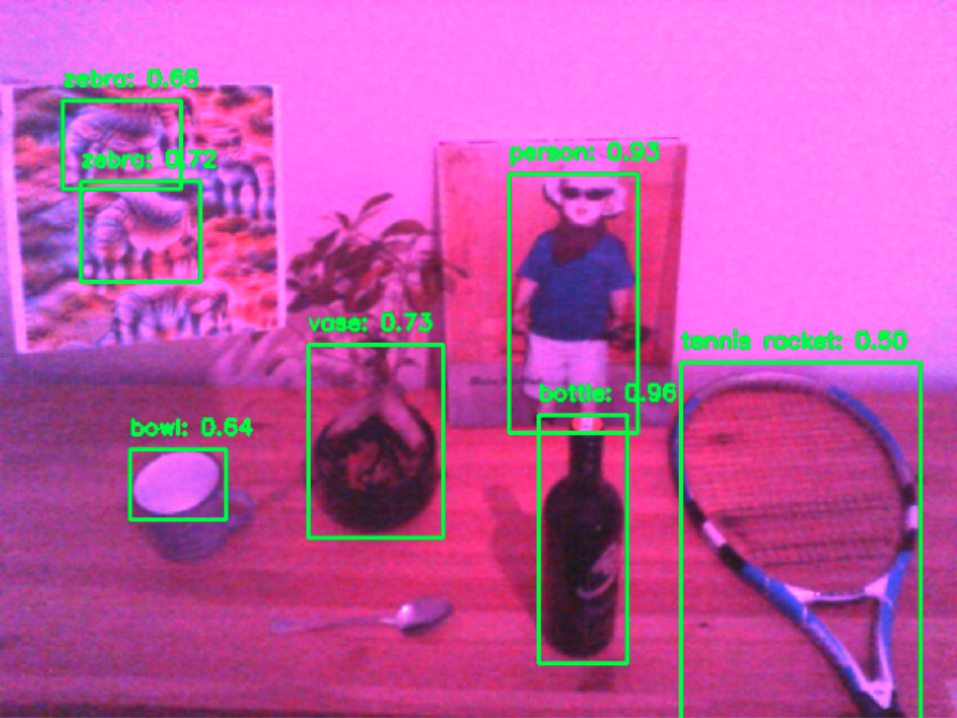}}
    
    \subfloat[Digitally changed hue
    \label{fig:digital_clean}]{\includegraphics[width=0.35\linewidth]{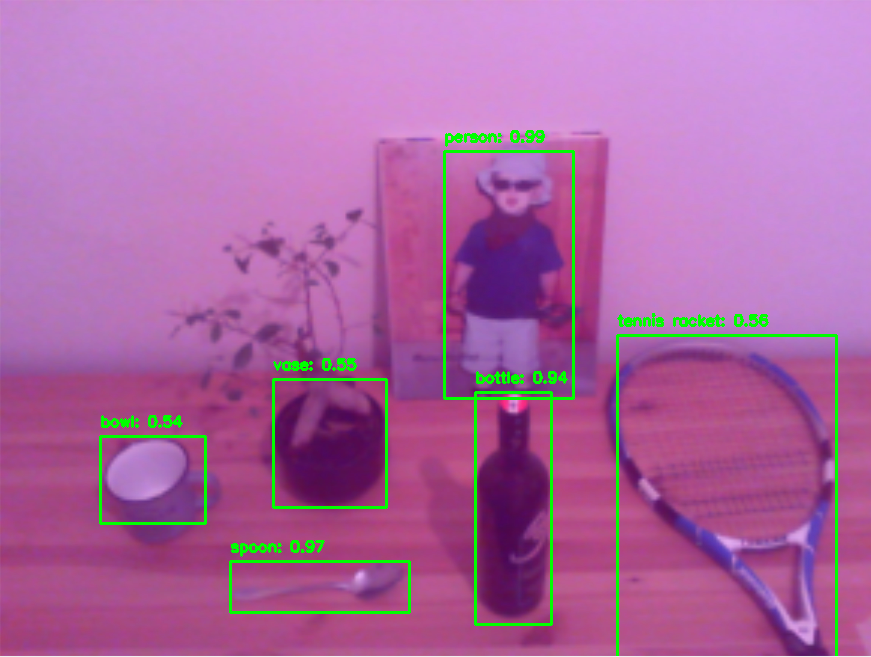}}
    \hspace*{10pt}
    \subfloat[As in (c) + patch = \color{blue}{effective}
    \label{fig:digital_patched}]{\includegraphics[width=0.35\linewidth]{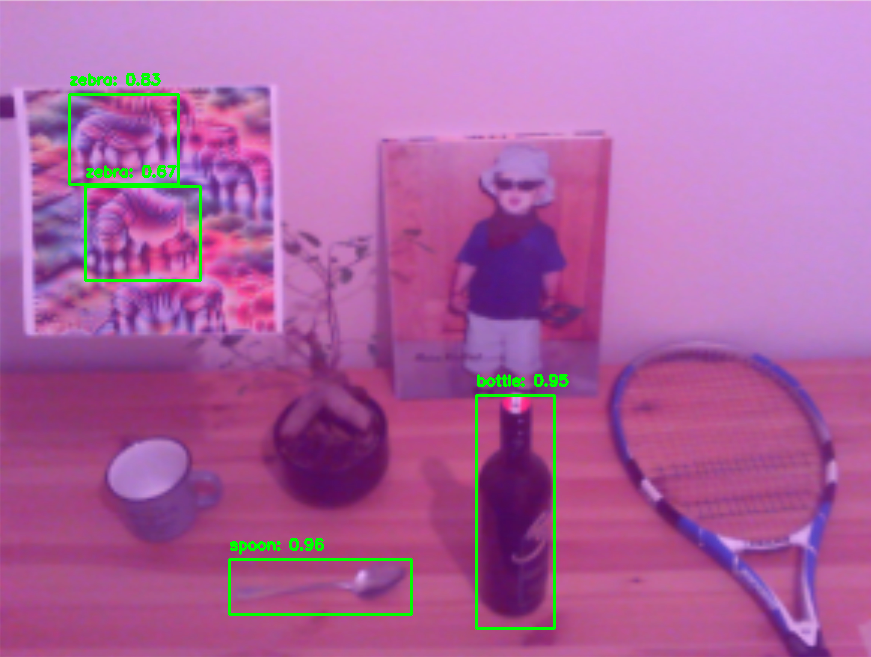}}
    \caption{Discrepancy between the hue transformation applied in the real world using a RGB light source (top) and digitally applied transformation using the best matching parameters got from a neural network (bottom).}
    \label{fig:firstfigure}
\end{figure}

This study focuses on the stability of adversarial patches in the physical world, aiming to understand how various environmental conditions and patch attributes affect their performance. Two types of adversarial patches were evaluated: a global patch designed to suppress all correct detections when placed anywhere in the scene, and a local patch targeting specific objects by partially overlapping them. The patches were tested on the same static scene with YOLOv3 and YOLOv5 serving as the detection networks, respectively. These versions were chosen due to the availability of pre-existing frameworks for generating adversarial patches tailored to them (~\citet{OnPhysicalAdversarial} and ~\citet{Patchv5}, respectively). The experimental setup included a controlled indoor environment with a standardized set of objects and lighting conditions. Performance was evaluated based on the mean average precision (mAP) for the global patch and detection confidence for the local patch. Key variables such as patch size, position, rotation, brightness, hue, blurriness and reduced color palette were systematically altered to assess their impact on patch efficacy. The experiments revealed significant dependencies between the performance of adversarial patches and these variables when applied in the digital and in the real worlds. While patch performance with respect to geometric transformation is consistent across both worlds, color transformations unveil substantial differences, which can't be easily matched, and indicating a gap between these both worlds. An example in \Figref{fig:firstfigure} shows the same scene, where the hue parameter was altered using a RGB light source (top) and digitally using the best parameters to match a physical change (bottom). YOLOv3 performance differs significantly when hue is changed physically and when hue is changed digitally. These findings highlight the sensitivity of adversarial patches to real-world conditions. This work is the extension of study conducted by~\citet{breakingtheillusion2024shack}.

\textbf{Contributions.}
Our study provides a comprehensive and systematic analysis of how adversarial patches perform under various physical-world conditions, including different lighting, patch sizes, and viewpoints. The findings underscore a significant impact of environmental conditions, such as lighting, on the effectiveness of adversarial patches. We show that the real-world effects are different from applying these transformations digitally using the best matching parameters. The study emphasizes the need for the development of advanced adversarial methods and improved defenses, contributing to the development of more resilient machine learning systems capable of withstanding real-world adversarial conditions.

\section{Related Work}

\textbf{Object detection models and their robustness.}
YOLO ~\citep{Terven_2023} is one of the most popular real-time object detection algorithms. Its high speed and good accuracy make it widely used in the community. The original YOLOv1 object detector was first released in 2016 by~\citet{redmon2016look}, and quickly became state-of-the-art. Over time, the algorithm was significantly improved, so different versions are now available~\citep{redmon2018yolov3, bochkovskiy2020yolov4, li2023yolov6, wang2022yolov7, yolov8_ultralytics}. In order to achieve the optimal performance of the object detection model, each YOLO version was trained with geometric (perspective change, scaling, translation, flipping, and rotation), color (HSV), and more advanced~\citep{wei2020amrnet, yun2019cutmix, 9578639, zhang2018mixup} augmentations. This made YOLO models resilient to challenging environments.

\textbf{Adversarial attacks and adversarial patches.}
Adversarial patches were first introduced in 2018 by~\citet{AdversarialPatch}, demonstrating the ability to mislead image classifiers using round stickers~\citep{wei2022adversarial}. This concept was extended to object detection in 2019~\citep{DPatch} with the Dpatch, although it was initially tested only in digital settings. Subsequent studies, such as the work by~\citet{OnPhysicalAdversarial}, adapted these patches for physical-world scenarios, highlighting challenges like maintaining attack efficacy across different environmental conditions. This type of attack is referred to as a \emph{global attack} in this study, as it targets the entire image. The paper proposes a way to generate a global patch attack, by maximizing the YOLO loss function. This global adversarial patch is primarily used for the experiments in this study.
A different approach for attacking object detectors is proposed by~\citet{Patchv5}. In this method, the patch must overlap the object that is intended to be hidden from the object detection network. This technique is referred to as \emph{local attack} because it targets individual objects within the scene.
The current research builds on these works by systematically evaluating the performance of both global and local adversarial patches.

\textbf{Defenses against adversarial patches.}
Adversarial attacks pose a substantial threat to object detection algorithms. Due to their disruptive potential of adversarial attacks, adversarial attacks have attracted considerable attention, with numerous researchers striving to devise innovative defense strategies~\citep{saukh2023,choi2022adversarial}. A common approach involves the localization and subsequent neutralization or removal of the adversarial patches.~\citet{jing2024padpatchagnosticdefenseadversarial} propose PAD - a patch-agnostic defense mechanism that combines semantic independence localization and spatial heterogeneity localization;~\citet{xu2022patchzerodefendingadversarialpatch} developed defense pipeline against white-box adversarial patches that zeros out
the patch region by repainting with mean pixel values;~\citet{naseer2018localgradientssmoothingdefense} proposed local gradient smoothing scheme that regulates gradients in the estimated noisy region of the image before inference;~\citet{scheurer2023detectiondefensespromiseadversarial} address defence against adversarial attacks in motion detection applications. In contrast to previous works, our carefully constructed experiments demonstrate that failure cases for existing adversarial patches can be deterministically constructed. These findings highlight the necessity for further research on more robust adversarial patches and stronger defense mechanisms.

\section{Methodology} 
\label{methodology}

\begin{figure}[t]
\centering
    \includegraphics[width=0.23\linewidth]{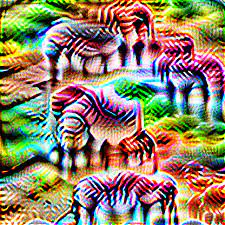}
    \hspace{10pt}
    \includegraphics[width=0.23\linewidth]{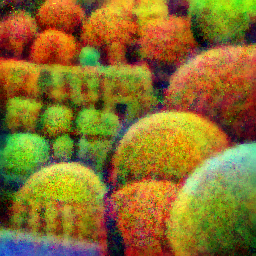}
    \caption{Global (left) and local (right) adversarial patches used in this work to attack YOLOv3 and YOLOv5.}
    \label{fig:YOLOvX_patch}
\end{figure}

\begin{figure*}[t]
\centering
    \subfloat[Physical patch left]{
    \includegraphics[width=0.235\textwidth]{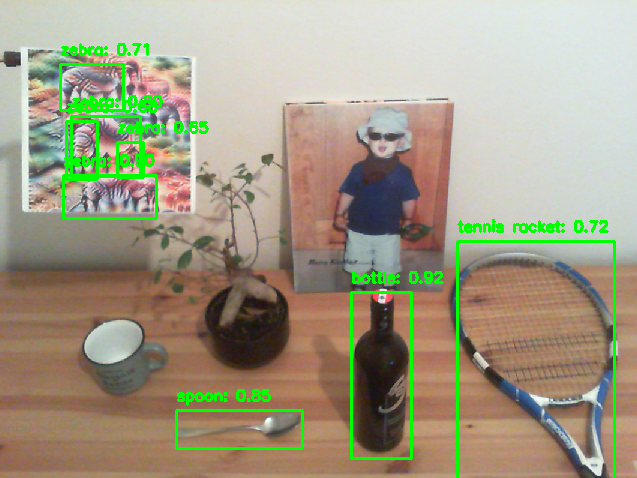}
    }
    \subfloat[Physical patch right]{
    \includegraphics[width=0.235\textwidth]{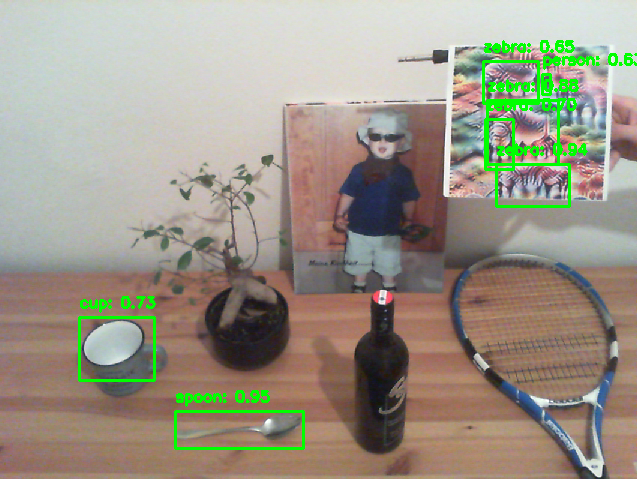}
    }
    \subfloat[Digital patch left]{
    \includegraphics[width=0.235\textwidth]{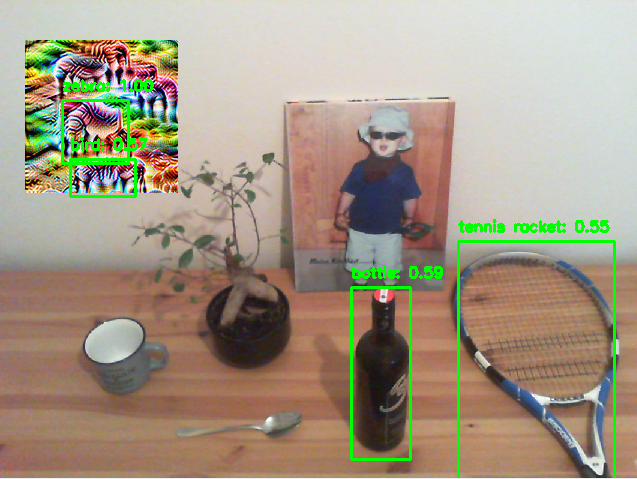}
    }
    \subfloat[Digital patch rotated]{
    \includegraphics[width=0.235\textwidth]{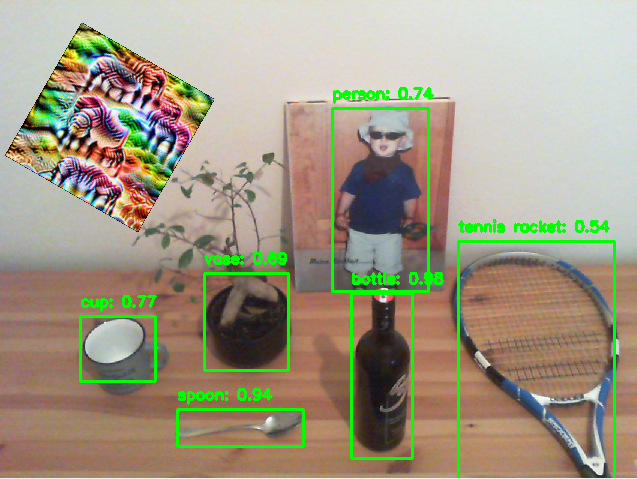}
    }

    \subfloat[Digital patch 12\% of the\\image]{
    \includegraphics[width=0.235\textwidth]{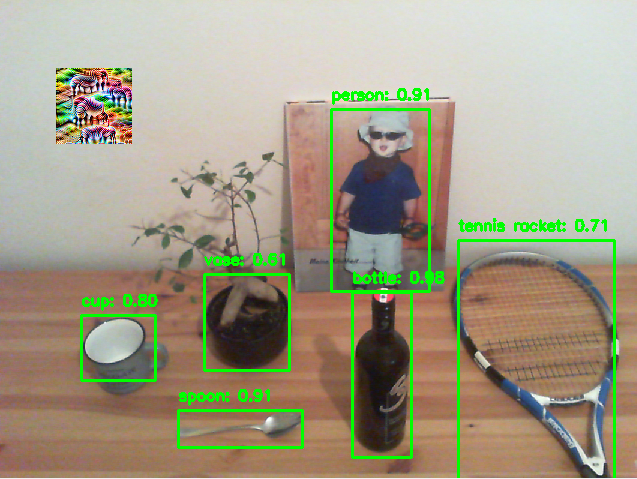}
    }
    \subfloat[Digital patch 30\% of the\\image]{
    \includegraphics[width=0.235\textwidth]{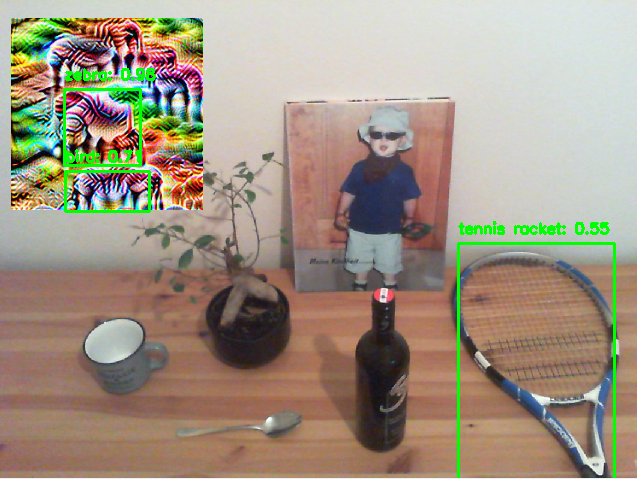}
    }
    \subfloat[Brightness 110 lux]{
    \includegraphics[width=0.235\textwidth]{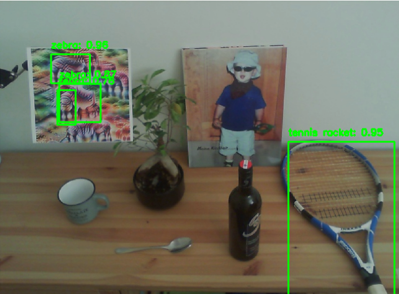}
    }
    \subfloat[Brightness 196 lux]{
    \includegraphics[width=0.235\textwidth]{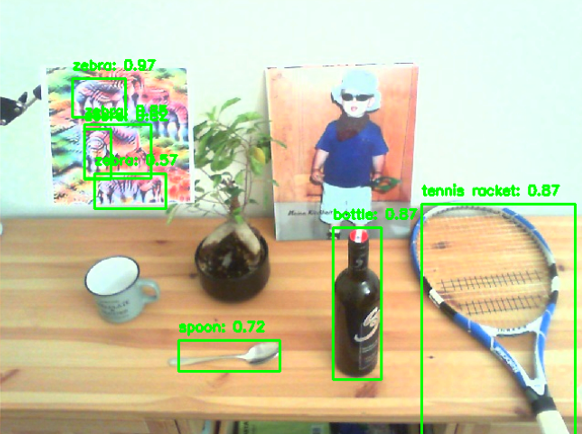}
    }
    \caption{Global patch performance under different conditions. (a)-(b): relocating physical patch shows its localized effects; (a)-(c): digital patch shows stronger performance when compared to a physical patch at the same position; (c)-(d): patch rotation mitigates its effect; (e)-(f): patch size correlates positively with its effectiveness; (g)-(h): the effect of environmental brightness is mixed.}
    \label{fig:hyp_dist}
\end{figure*}

\textbf{Adversarial patch generation.}
The attack patches used in this work were generated by~\citet{OnPhysicalAdversarial} and~\citet{Patchv5}, using variants of local and global projected gradient descent running the following optimization: 
\begin{equation}
\arg \max_{\delta} \mathbb{E}_{(x,y) \sim \mathcal{D}, t \sim \mathcal{T}} [J(h_{\theta}(A(\delta, x, t)), y)],
\end{equation}
\noindent where $\mathcal{D}$ is the distribution over samples, and $A$ is the patch application function. The function $A$ applies a transformation $\delta$ with parameters $t \in \mathcal{T}$ to the patch during training to ensure robust patch performance (for example, the global patch was trained with rotation augmentation). The patch is then integrated into the image $x$ at a desired location. The optimization is solved essentially by using gradient descent~\citep{OnPhysicalAdversarial,DPatch,Patchv5}.

\textbf{Global and local patches.}
This study evaluates two types of patches: a global patch and a local patch. The global patch, designed to attract the attention of the object detection network and suppress correct detections, can be placed anywhere in the image and was generated as described by~\citet{OnPhysicalAdversarial} using YOLOv3~\citep{redmon2018yolov3}. The local patch, which must overlap the target object, was generated according to~\citet{Patchv5} using YOLOv5~\citep{yolov8_ultralytics}. Both patch generation processes used the COCO2014 dataset~\citep{COCO2014}. YOLOv3~\citep{redmon2018yolov3} was used to test the global patch, and YOLOv5~\citep{yolov8_ultralytics} was used for the local patch, as these versions were originally used to generate the patches.
The generated patches (\Figref{fig:YOLOvX_patch}) are specific to their respective detection networks and are not transferable.
All physical patches were printed on regular paper using a standard printer.

\textbf{Hypothesis.}
Following recent findings that adversarial patches may fail in the physical world~\citep{hartnett2022empirical}, we conducted a dedicated set of experiments to better understand these failure cases. To achieve this, we (1) carefully constructed our experiments, and (2) investigated the differences between the effects observed in the physical world and their reproducibility through digital transformations. We used sensors, and two cameras operating in well-documented modes to run reproducible real-world experiments. Our main hypothesis is that failure cases of adversarial patches in the physical world in general differ significantly from similar experiments conducted digitally, \ie by embedding the patch into an image and transforming the result using the same parameters as measured physically, or the best matching parameters computed by an optimization algorithm. We present this analysis next.

\section{Discovering Vulnerabilities of Adversarial Patches}
\label{sec:vulnerabilities}

\subsection{Experimental setup}
\textbf{Controlled real-world environment.}
We evaluated the performance of adversarial patches by conducting physical attacks in a controlled indoor setting and attempting to reproduce them digitally for comparison. This controlled environment allowed us to easily adjust testing conditions. Lighting control was managed using an IKEA\textsuperscript{\textregistered} Tradfri LED1924G9 RGB light source. We primarily used the Microsoft LifeCam HD-3000 camera, which records 720p HD videos at up to 30 fps. To ensure results were not camera-specific, we repeated experiments involving brightness and hue with the Ausdom AF640 camera, which records 1080p HD videos at up to 30 fps.
Our scene setup is shown in \Figref{fig:firstfigure}. The test included a bottle, cup, small potted plant in a vase, tennis racket, spoon, and a picture of a person. 
Occasionally, a dining table was detected with low confidence but excluded from consideration due to inconsistency.

\textbf{Evaluation metrics.}
To evaluate the performance of the global patch, we primarily use the mean average precision (mAP) as the metric. mAP is calculated by generating precision-recall curves and then determining the area under these curves. This metric provides insights into the overall performance of an object detection system. In this study, a lower mAP of the detection algorithm indicates better performance of the patch in suppressing detections. For the local patch, we measure performance by the detection confidence of the targeted object. The detection confidence assigned to each detection describes the confidence or probability of a detected object belonging to a particular class. Lower detection confidence signifies higher patch effectiveness.

\subsection{Experimental variables}
Following setups described in the literature~\citep{Chen_2019, eykholt2018robust, thys2019fooling, braunegg2020apricot}, we varied key parameters that can also easily change in uncontrolled real-world settings: (1) geometry (patch size, observation angle, distance to the target), (2) color transformations (scene brightness and hue), and (3) information reduction (blurriness and limiting the number of colors in an image).

\textbf{Geometric transformations.} 
We first experimented with different patch sizes. For global attacks, patch sizes ranged from 10\% to 30\% of the image width to avoid object overlap. For local attacks, we tested patch sizes varied from 4cm x 4cm to 16cm x 16cm.
For all other experiments (geometric, color or information reduction), by default, we used a 25\% image width-sized patch for global attacks and the smallest patch that significantly reduced object detection confidence for local attacks (11cm x 11cm for the tennis racket, 7cm x 7cm for the other objects). We explore patch rotations up to 90$^\circ$ around X, Y, and Z axes (see Figure~\ref{fig:rot_expl}).
We also explore the impact of the global patch position within the scene on its detection suppression ability depending on the distance from the target. 

\textbf{Color transformations.}
Ambient brightness was varied from 4 to 61 lux (measured with a light sensor). With automatic camera exposure, there is a trade-off between the image brightness and noise. To mitigate this, we fixed the exposure time, thereby enabling overexposure -- a common problem in real-world applications like autonomous driving~\citep{8569692}. The camera exposure was calibrated to produce a uniform, naturally looking image at a measured brightness of 15 nits. However, a discrepancy persists between the measured lux in the room and the illuminance calculated from the scene image. 
To address this issue and facilitate comparisons with our digital experiments, we performed an illuminance scale correction based on the calculated image illuminace. This adjustment shifted the real-world range of 4 to 61 lux to an approximate range of 68 to 243 lux as measured in the images.  
We varied the hue values of the scene with an IKEA Tradfri LED1924G9 RGB light source (see Figure~\ref{fig:firstfigure} for an example). Note that this light source also influences the other colour properties of the environment (\eg brightness). 

\textbf{Information reduction.}
We conducted a series of information reduction experiments in the digital domain to analyze the performance of a low-quality patch due to possible camera or printing effects. A low-pass filter is often used in digital image processing domain to smooth the image, soften the sharp regions, and remove the noise while preserving important image features. We varied the filter size from 0 to 500. 
Color reduction filtering aims to enhance image compression, optimize storage efficiency, and decrease computational complexity in image analysis tasks by reducing the number of distinct colors in an image while maintaining its essential visual features. The number of reduced colors was varied from 2 to 600, whereas a natural-looking image of a scene contains more than 50'000 different colors. 

\begin{figure}[t]
    \includegraphics[width=0.50\linewidth]{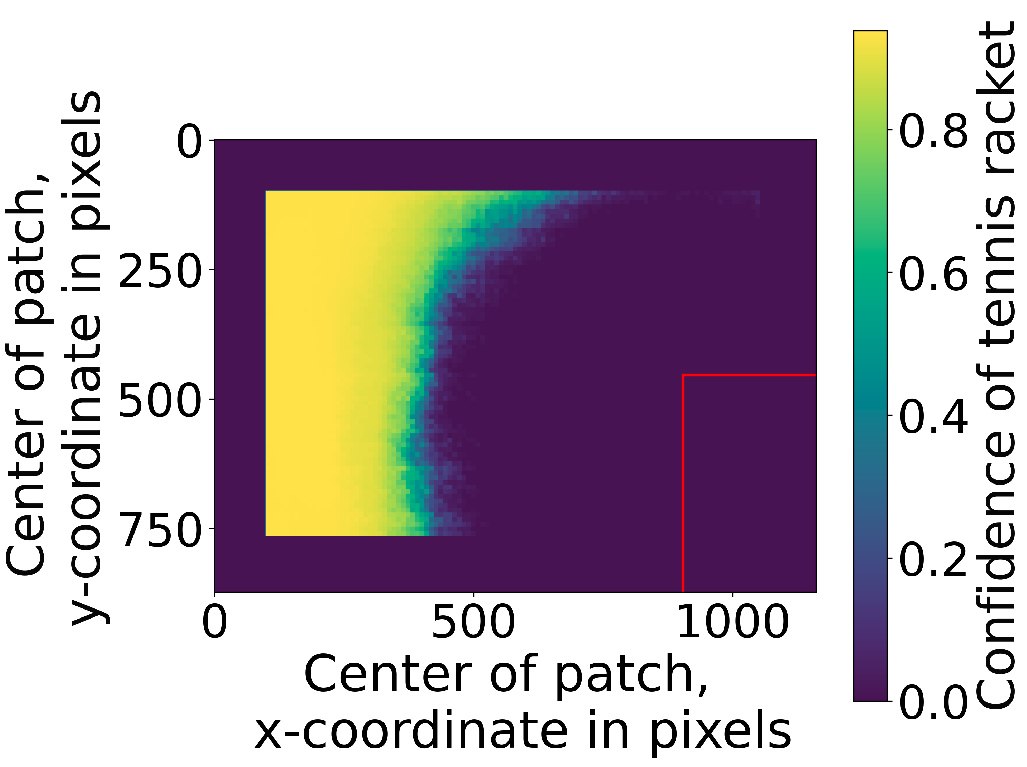}
    \hspace{10pt}
    \includegraphics[width=0.40\linewidth]{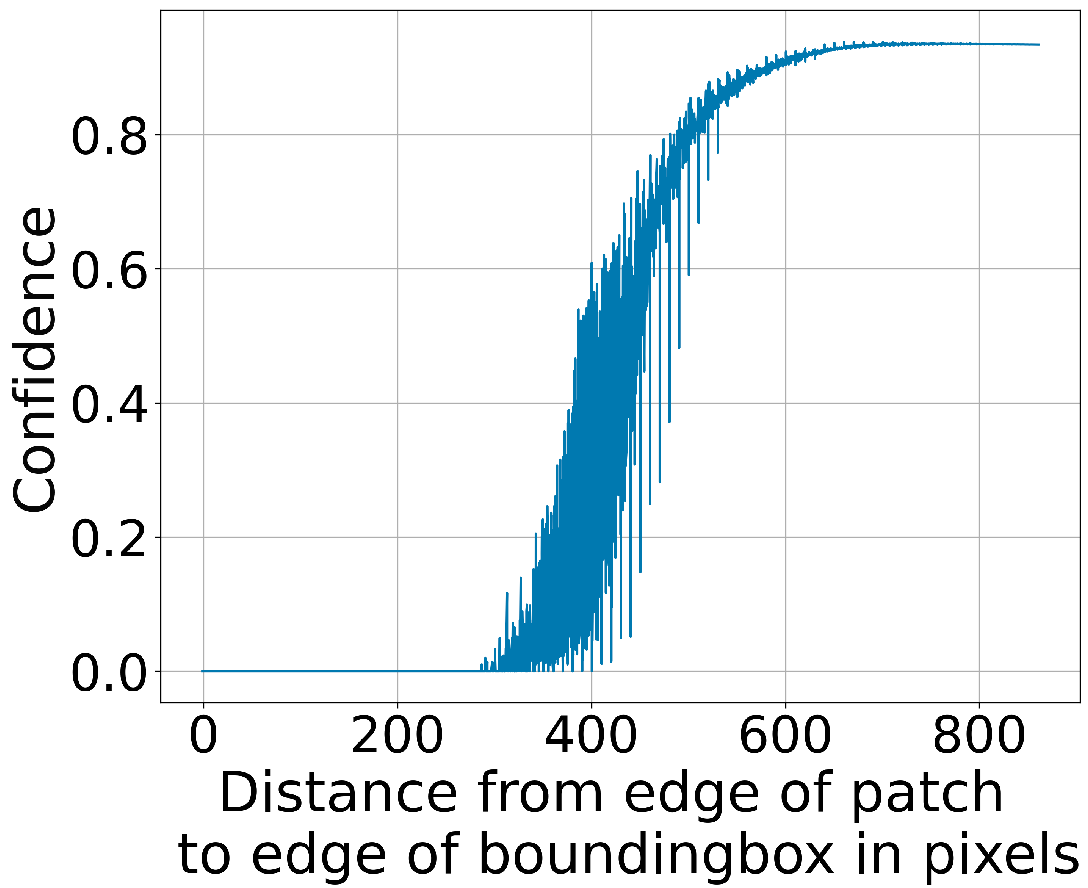}
    \caption{{Detection confidence over patch position (left), and distance between the edge of the patch and the bounding box (right) for "tennis racket".}}
    \label{fig:conf_tennis_racket}
\end{figure}

\subsection{Exploration based on a fixed indoor scene} \label{sec:exploration}
To get a better understanding of the stability of the attacks in the real world, we run comprehensives experiments and attempt to reproduce the results in the digital world. The observations below relate to the global patch, but most our findings also apply to the local patch, which is discussed in the appendix.

\textbf{Distance dependence.} \label{distance_dependence}
The first stability issue is the patch's effectiveness depending on its distance from the object. Experiments show an attack is successful only if the patch is within a reasonable distance from the target. Comparing \Figref{fig:hyp_dist}(a) and \Figref{fig:hyp_dist}(b) confirms the patch is more effective when closer to the object. The original paper~\citep{OnPhysicalAdversarial} claims that while the patch is somewhat location-invariant, its influence weakens with distance. We further investigated the effect by digitally inserting the global patch at various positions. \Figref{fig:conf_tennis_racket}(left) displays tennis racket detection confidence, with the x and y positions indicating patch placement and confidence levels. The red rectangle is the ground truth bounding box. Results show the patch must be within a certain radius, dependent on the size of the patch, resolution of the scene, and the object itself, to suppress detection effectively. \Figref{fig:conf_tennis_racket}(right) illustrates detection confidence relative to the distance from the patch's edge to the bounding box edge, showing the patch loses its adversarial properties when positioned too far away from the object ($\sim$400px).

\begin{figure}[t]
    \centering
    \includegraphics[width=0.40\linewidth]{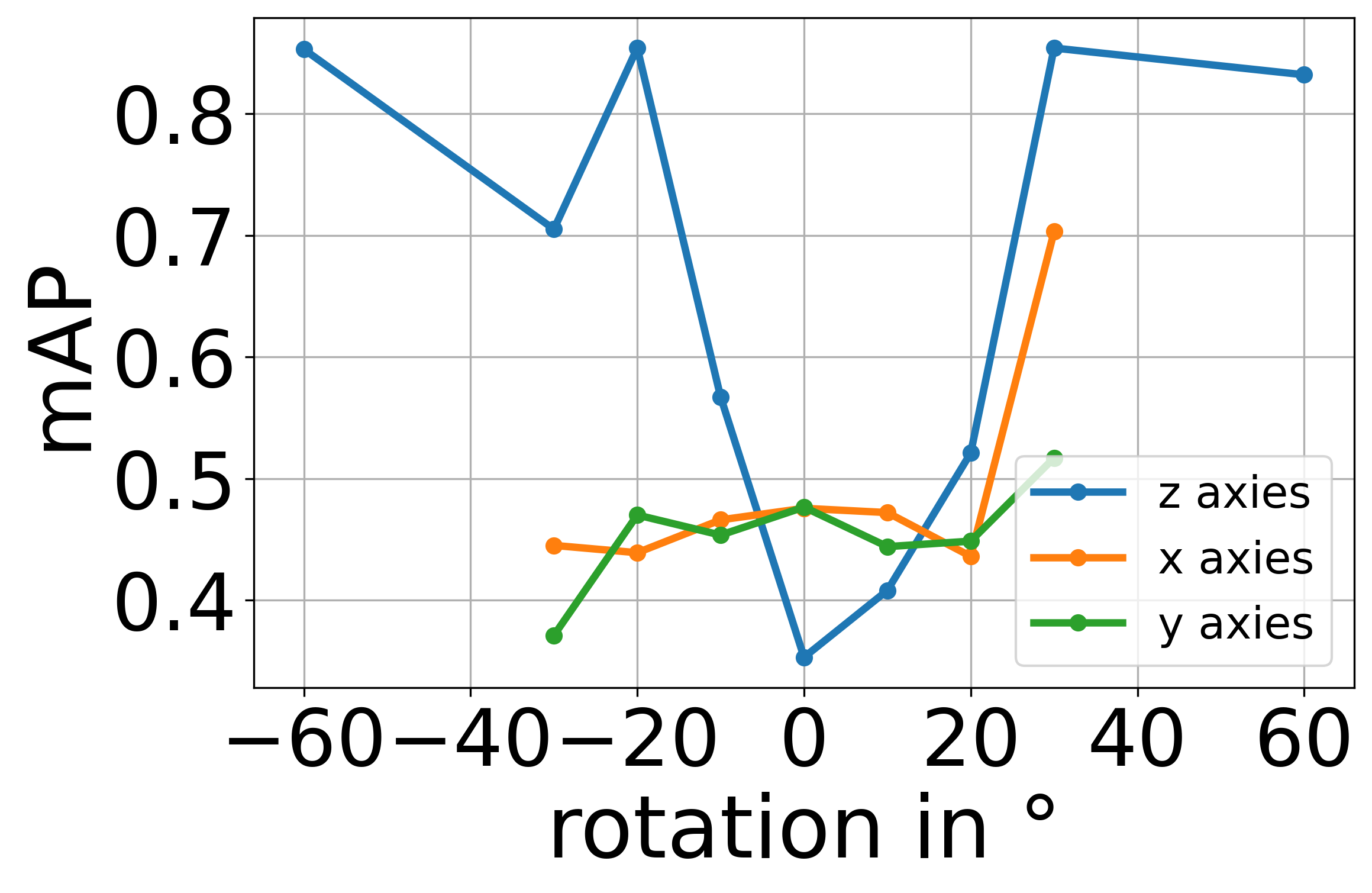}
    \hspace{10pt}
    \includegraphics[width=0.40\linewidth]{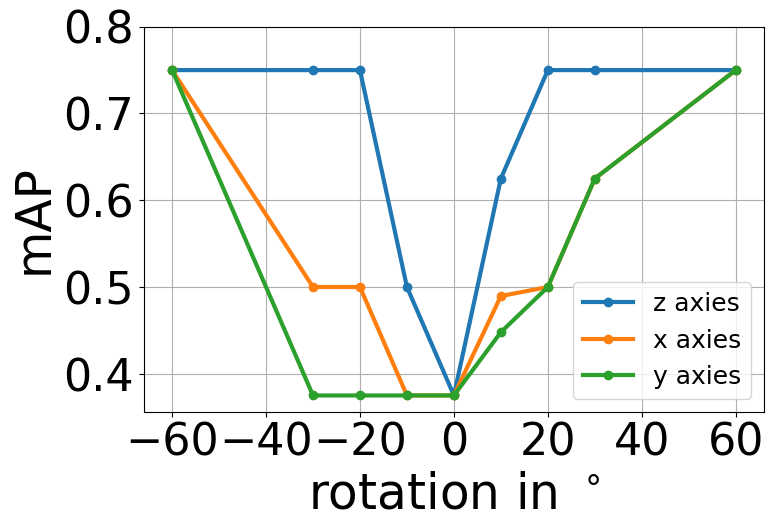}
    \caption{Mean average precision for patch rotations around all axes: physical experiment (left) and digital experiment (right). The patch sizes across domains were aligned.}
    \label{fig:mAP_o_angle}
\end{figure}

\textbf{Rotation dependence.}
Due to the nature of physical experiments, the position of the patch relative to the camera is crucial in an adversarial attack. \Figref{fig:hyp_dist}(d) illustrates a large rotation around the z-axis, resulting in a significant reduction in adversarial performance compared to \Figref{fig:hyp_dist}(c). 
Rotation is often included in training neural networks for computer vision as a data augmentation strategy~\citep{cao2023mitigating}. Rotation transformation was also applied in the global patch generation software~\citep{Patchv5,zhang2023puerh}. Consequently, we expect the patch to exhibit some robustness to rotations. \Figref{fig:mAP_o_angle} shows the mAP over rotation angles across the three axes in the real world (left) and digitally (right). The patch sizes were aligned across these two settings. In both cases, the patch shows robustness to rotations around x and y axes within $\pm$~40$^\circ$, yet loses its adversarial properties for rotations around z axis larger than 20$^\circ$. Adversarial effects of the patch in the digital domain is stronger.

\begin{figure}[t]
    \centering
    \begin{minipage}{.40\linewidth}
        \centering
        \includegraphics[width=\linewidth]{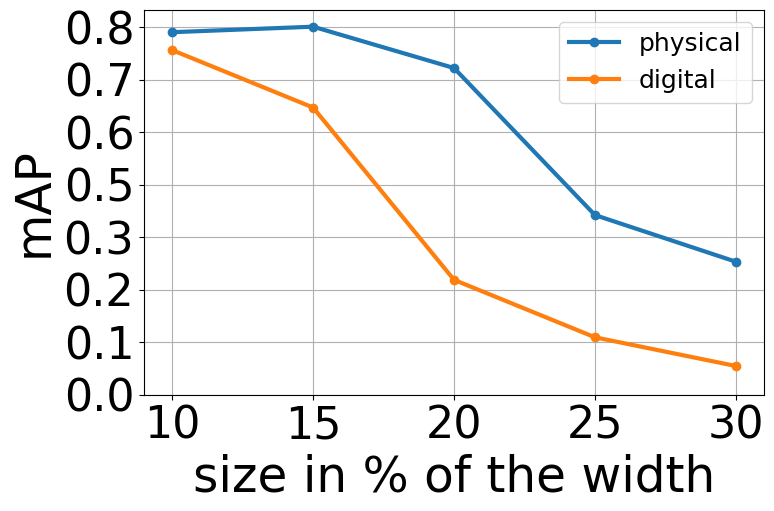}
        \caption{Mean average precision over patch size.}
        \label{fig:mAP_o_size}
    \end{minipage} 
    \hspace{10pt}
    \begin{minipage}{.40\linewidth}
        \centering
        \includegraphics[width=\linewidth]{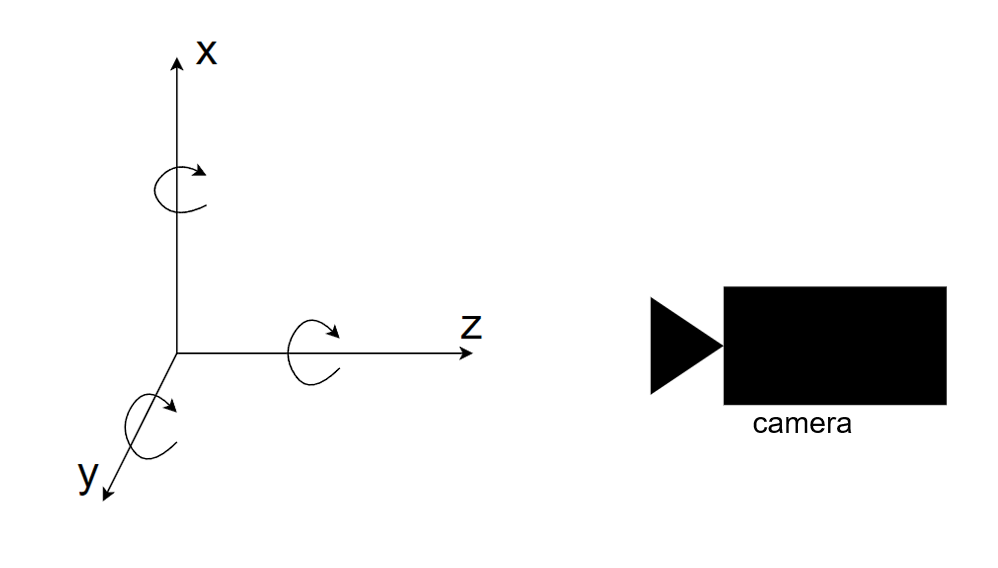}
        \caption{Axes placed in space relative to the camera.}
        \label{fig:rot_expl}
    \end{minipage}
\end{figure}

\textbf{Size dependence.}
The patch size positively correlates with its effectiveness. \Figref{fig:hyp_dist}(e) and \Figref{fig:hyp_dist}(f) show a significant difference in performance when scaling a global patch from 12\% to 30\% of the image. This trend is confirmed digitally, as \Figref{fig:mAP_o_size}(left) shows larger patches suppress more detections than smaller ones, with the effect being stronger in the digital domain.

\begin{figure}[t]
    \centering
    \includegraphics[width=0.40\linewidth]{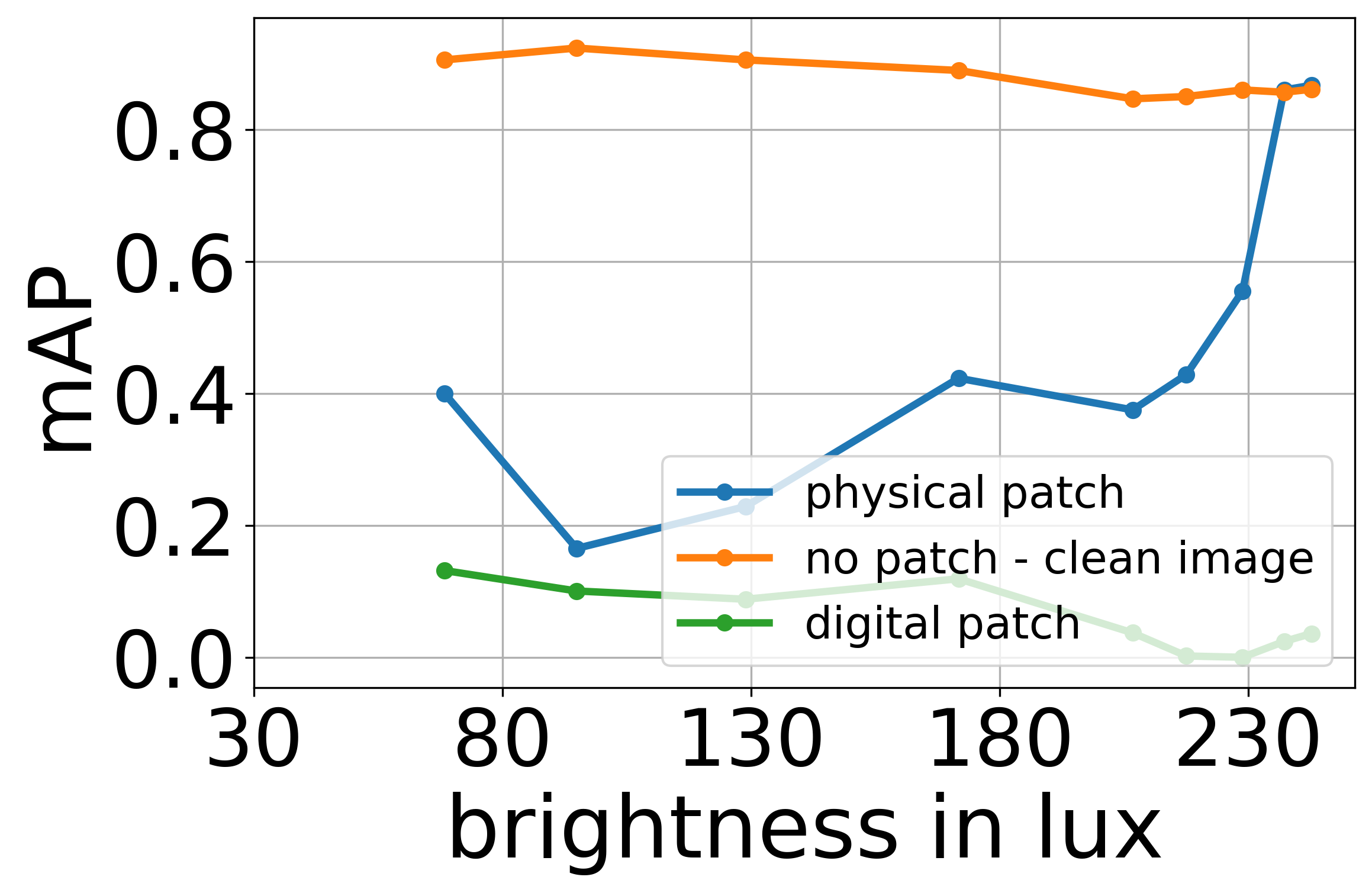}
    \hspace{10pt}
    \includegraphics[width=0.40\linewidth]{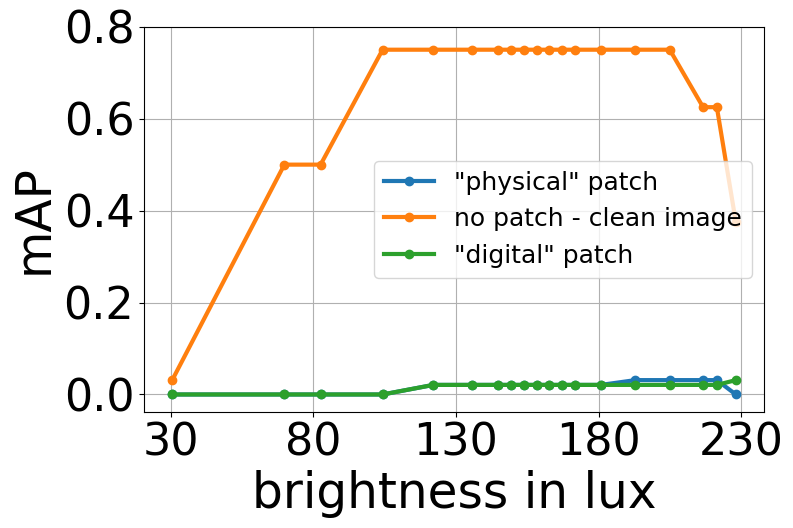}
    \caption{Mean average precision over scene brightness: physical experiment (left) and digital experiment (right).}
    \label{fig:mAP_brightness}
\end{figure}

\textbf{Brightness dependence.} \label{brightness_dependence}
Another issue with attack stability becomes apparent when lighting conditions change. If the camera's exposure time is set to automatic, there is a trade-off between image brightness and noise. Setting this to manual removes this dynamic adaptation and enables us to emulate overexposure. The exposure time is fixed and calibrated to produce a uniform and naturally looking image for the baseline of the experiment at 15 nits. \Figref{fig:hyp_dist}(g) shows an example of patch performance at reduced brightness, while \Figref{fig:hyp_dist}(h) shows an example at increased brightness. Here, the adversarial patch loses effectiveness if the image becomes too bright and starts clipping. Conversely, decreasing brightness does not significantly impact the patch's performance. \Figref{fig:mAP_brightness} clarifies how patch performance is affected by brightness. The digital patch shows consistent performance unaffected by lighting changes over the whole range of values. The physical patch, however, looses its effectiveness with higher brightness when clipping occurs, matching the mAP of a clean image without a patch.

\begin{figure}[t]
    \centering
    \includegraphics[width=0.40\linewidth]{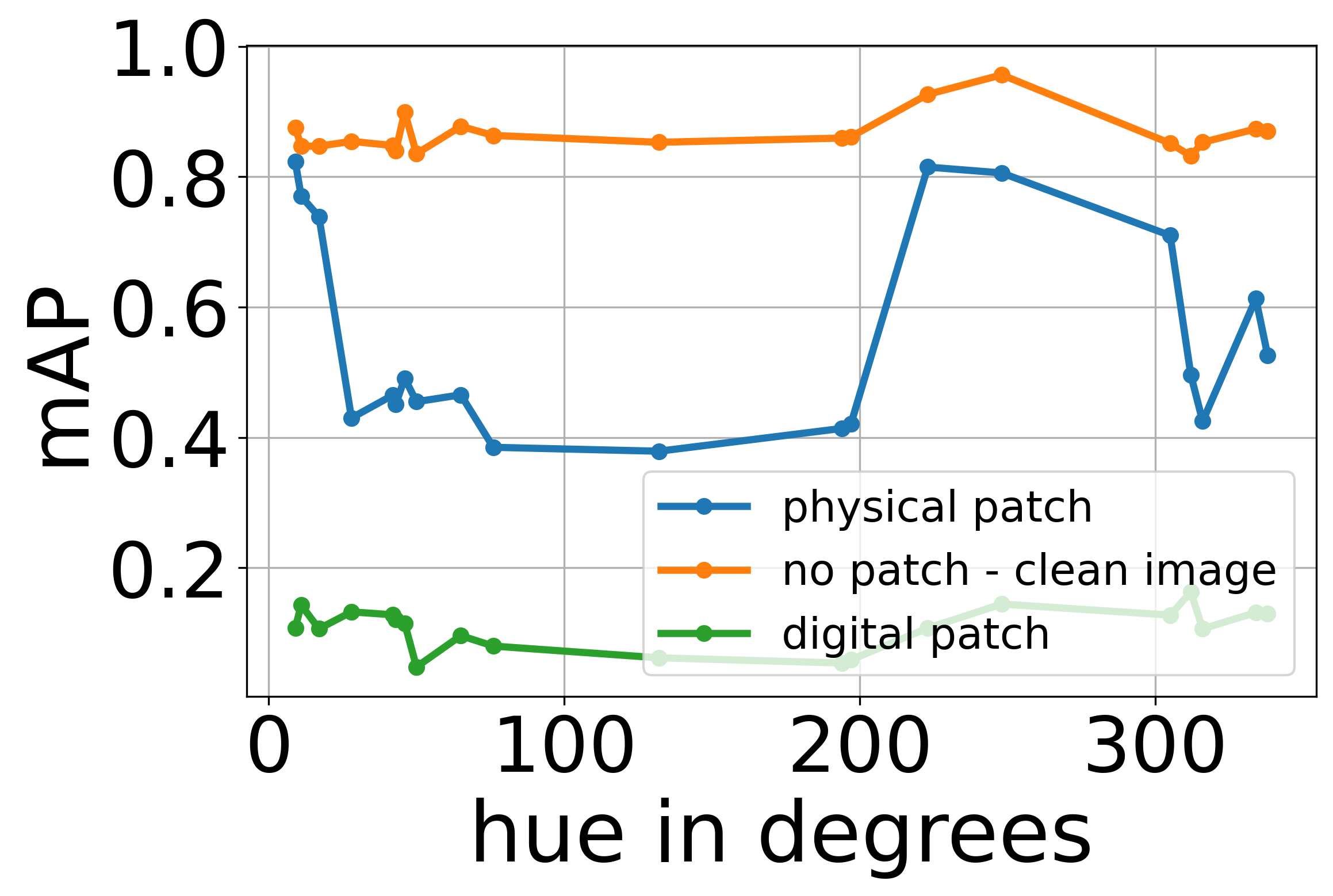}
    \hspace{10pt}
    \includegraphics[width=0.40\linewidth]{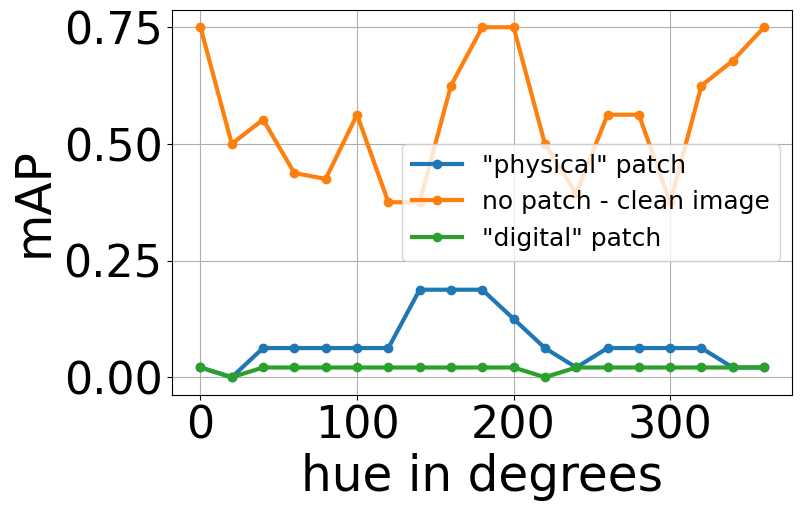}
    \caption{Mean average precision over scene hue: physical experiment (left) and digital experiment (right).}
    \label{fig:mAP_hue}
\end{figure}

\begin{figure}[t]
    \centering
    \includegraphics[width=0.40\linewidth]{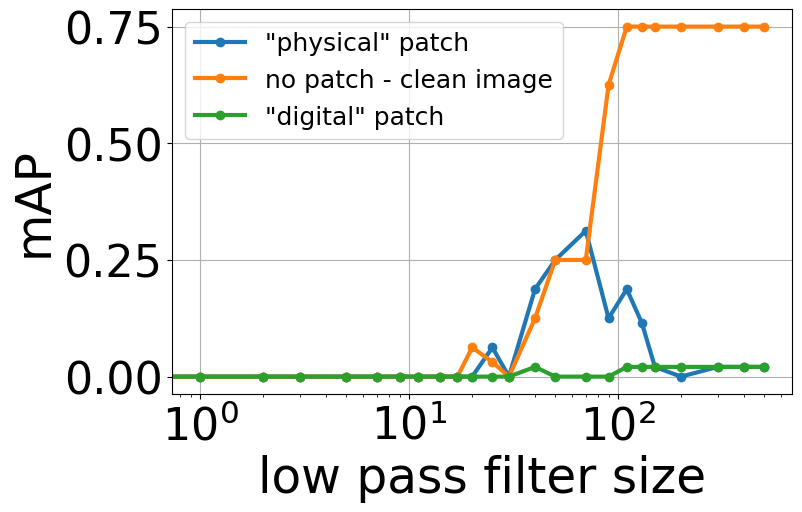}
    \hspace{10pt}
    \includegraphics[width=0.40\linewidth]{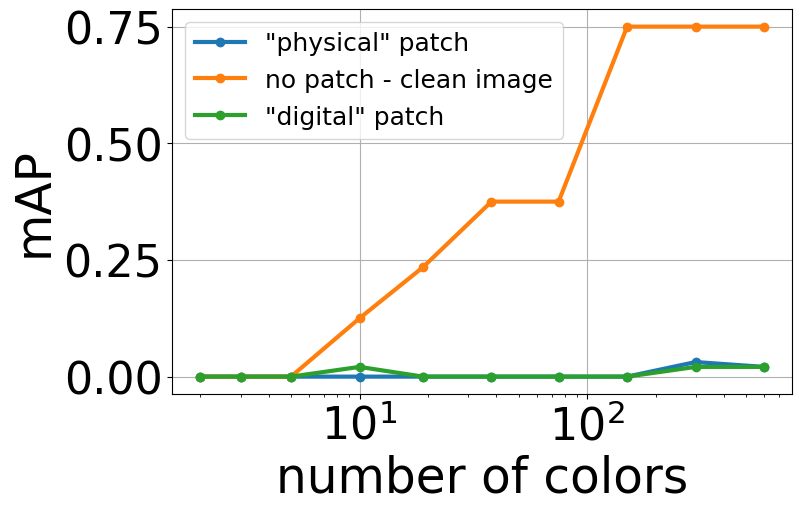}
    \caption{Mean average precision over low pass filter size (left) and number of colors in the image (right).}
    \label{fig:mAP_lowPassFilter_colorReduction_digital}
\end{figure}

\textbf{Hue dependence.}
Here we investigate patch performance when changing hue lightning. \Figref{fig:firstfigure} provides examples of images with physically and digitally altered hue and the corresponding detection results. YOLO achieves a mAP of 0.4 in real-world (\Figref{fig:firstfigure}(b)), and a mAP of 0.14 in the digital world (\Figref{fig:firstfigure}(d)), resulting in approximately 64\% discrepancy in patch performance. In the real world, we used a RGB light source to change hue. To get the hue value of the light source, the value reported in the companion app (IKEA\textsuperscript{®} Home smart 1) to the LED was used. To digitally replicate physical world scene images as accurately as possible, \ie to digitally generate the scene images that are the best match to the  physical scene images, we train a small neural network.
Our neural network outputs suggest that color transformations alone do not fully capture the changes introduced by an additional hue light source. To support this, we present the performance of the patch across the complete hue range in \Figref{fig:mAP_hue}. \Figref{fig:mAP_hue} (left) shows the results from physical experiments, where the hue value is reported by the additional light source app. \Figref{fig:mAP_hue} (right) displays results from digital experiments, where only the hue value of the image is digitally altered. Notably, YOLO performs consistently across all hue values in the physical world, whereas digital experiments exhibit some fluctuations in YOLO performance. Additionally, in the physical world, there is a clear range in the hue spectrum, between 200 and 300 degrees, where the patch fails to perform effectively. In contrast, digital experiments show only slight disturbances in patch performance between 150 and 200 degrees.

\textbf{Information reduction.}
This set of experiments demonstrates that the patch requires a certain amount of information to become effective. In the low-pass filter experiments shown in \Figref{fig:mAP_lowPassFilter_colorReduction_digital}(left), the detection efficiency for the scene without a patch and the scene with the simulated physical patch (labeled as "physical" patch) align closely up to a certain point -- \ie the presence of the patch has no influence on the detections. 
Beyond this point, the presence of the patch decreases detection efficiency.
In the color reduction experiments shown in \Figref{fig:mAP_lowPassFilter_colorReduction_digital}(right), there is almost perfect alignment between the scene with the simulated physical patch and the scene with the digital patch. 
The detection algorithm also requires a certain amount of color details to achieve its full detection potential. While performing these experiments,  we observed that the colors on the patch are quite diverse and mixed. Consequently, when a color reduction filter is applied, the patch retains a relatively large number of colors compared to the real-world environment. This allows the patch to maintain the maximum of it's efficiency and remain as efficient as the original digital patch.

\section{Discussion, Limitations, and Outlook}

\textbf{Discussion.}
Our experiments reveal significant dependency of patch effectiveness on environmental variables, such as patch size, position, rotation, brightness, and hue, highlighting the challenges in maintaining attack efficacy in real-world scenarios. While some failure cases are intuitive (\eg a positive correlation between the patch size and its effectiveness), many of our observations are not. Despite our best efforts to match the transformation parameters in the real world and in the digital domain, patch performance discrepancies remain, leaving many questions open. An interested reader is invited to check the outcome of our experiments for the local patch in the Appendix.

Lighting conditions, particularly brightness and hue, were found to be critical for patch performance. Changes in brightness, whether from natural or artificial lighting, can alter the patch's appearance and its interaction with the detection model. Overexposure or underexposure can reduce patch's effectiveness, as observed in our experiments. Variations in hue from different light sources also impact the patch's ability to disrupt detection. 
In the physical world, light interacts with different materials in complex ways, influencing the appearance of objects. Geometrical and physical optics, including reflection, scattering, interference, and absorption contribute to the perception of color~\citep{Tilley2020}. The color of an object is primarily determined by the energy of the light wave and material's properties - the energy of the incident wave, 
surface roughness and texture. The light that is leaving the object is what is seen. Modeling these interactions digitally presents significant challenges, as highlighted by~\citet{Musbach2013}.

The impact of training data on performance of patches tailored for the specific detection algorithm is a critical consideration. 
The COCO dataset's 80 classes are highly imbalanced. \textit{Person} is the most frequent class, occurring over 250'000 times, while \textit{bottle} and \textit{cup} appear approximately 25'000 times each, and \textit{tennis racket} only about 5'000 times. Our experiments showed that concealing the tennis racket is challenging, often requiring larger and closer patches, likely due to the skewed distribution of training data. 
Furthermore, patch behaviour under varying hue values may be attributed to the absence of red and violet images in the COCO dataset - most images are yellow and green hue.

\textbf{Limitations.}
Our study has several limitations. Firstly, the experiments were conducted in a controlled indoor environment, which does not capture the variability of a real outdoor settings. Factors like weather, outdoor lighting, and movement dynamics were not considered. The physical patches were printed on standard paper and evaluated under specific conditions, without exploring variations in patch materials and printing quality. 
Secondly, the study focused on a limited set of objects and scenes. While the selected objects provided a good baseline, a more diverse set of objects and scenes could reveal more insights. The experiments were limited to two types of adversarial patches and specific versions of the YOLO network. Exploring other types of patches and different object detection models could further generalize the findings.

\textbf{Future Work.}
The findings call for a deeper understanding of the interplay between adversarial strategies and environmental factors.
Future research should focus on developing more sophisticated adversarial methods that can adapt to changing conditions, and on improving the robustness of detection models to withstand such attacks, thereby enhancing the security and reliability of machine learning applications in real-world settings. While adversarial patches are static, object detection models can take advantage of on-device adaptation and reconfiguration to improve their resilience to adversarial attacks, \eg~\citet{saukh2023,wang2024subspaceconfigurablenetworks}.

\section*{Acknowledgments}

We would like to thank Dr Ramona Marfievici for her valuable guidance and feedback as the shepherd of our paper for Workshop on Enabling Machine Learning Operations for next-Gen Embedded
Wireless Networked Devices (EMERGE 2024). Her comments and suggestions greatly improved the quality of this work.

\clearpage
\bibliography{arxiv_cites.bib}

\newpage
\appendix
\section*{Appendix}
\section{Parameter Matching}
\label{appx:yolo_training}

\subsection{Geometric Transformations}
For geometric transformations, such as patch size and observation angle, we directly matched the parameters between the real-world and digital experiments. This alignment is feasible because, in both settings, patch size and observation angle are measured relative to the scene as captured by the camera.

\subsection{Color Transformations}
Matching parameters for color transformations proved to be more complex. 

\textbf{Brightness Transformations.}
For brightness transformations, ambient brightness was varied from 4 to 61 lux, as measured with a light sensor. However, due to various factors such as the camera’s exposure settings, light distribution in the room, and the light sensor’s position, the measured illumination did not correspond to the calculated illumination from the scene images. To match the physical and digital experiment parameters, we performed an illuminance scale correction based on the calculated image illuminance from the real-world experiment images. This adjustment shifted the real-world range of 4 to 61 lux to an approximate range of 68 to 243 lux.

\textbf{Hue Transformations.}
For hue transformations, we initially conducted a direct matching of the real-world and digital parameters and the results are shown in \Figref{fig:mAP_hue}. The results prompted further investigation, leading us to conclude that a hue light source in the real world affects not only the hue value that can be seen digitally, but also other colour properties of the environment visible digitally. To accurately replicate the physical world in the digital domain,~\ie to digitally generate the scene images that are the best match to their physical counterparts, we train a convolutional neural network. 

\begin{figure}[h]
    \centering
    \includegraphics[width=1\linewidth]{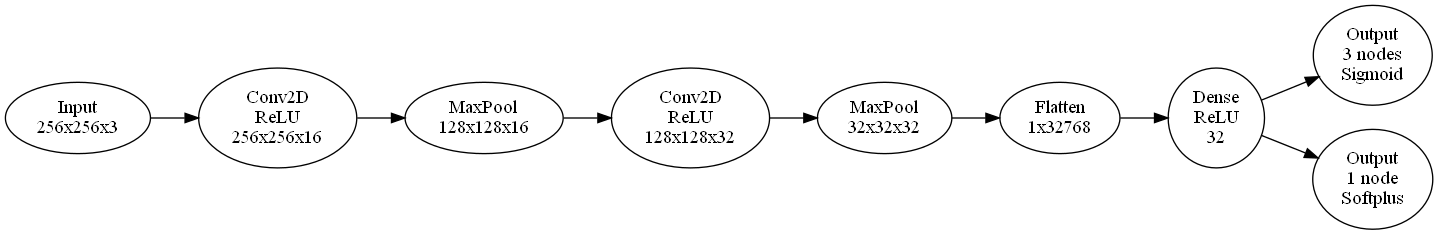}
    \caption{Neural network architecture.}
    \label{fig:nn_architecture}
\end{figure}

The architecture of the neural network is illustrated in \Figref{fig:nn_architecture}. It processes scene images (targets) of size 256x256x3 in the input layer. This is followed by two pairs of convolutional layers, each with ReLU activation and Max-Pooling layers. Subsequently, there is a fully connected layers with ReLU activation. The output layer splits into two branches: one with 3 nodes using the Sigmoid activation function for transformation parameters that are within a specific range (brightness factor, contrast factor, hue factor), and the another with 1 node using the Softplus activation function for transformation parameters that are positive values (saturation factor). 

We utilized images of our scene, varying the hue and ambient brightness using an IKEA Tradfri LED1924G9 RGB light source and an additional light source. The dataset comprises 28 images: 23 for training, 4 for validation and 1 for testing. The input images are resized to 256x256 pixels and standardized using the target data's mean and standard deviation, to being fed into the network. The output parameters are then applied digitally to the baseline scene image, taken under natural lighting conditions (without additional hue light or other light sources), to approximate the target image. We used mean squared error loss function and the Adam optimizer with a learning rate of 0.0001.  The model was trained with a batch size of 4 over 500 epochs. To prevent overfitting, a dropout rate of 0.5 was applied after fully connected layers.

This task proved to be challenging. The addition of a hue light in the physical world significantly alters the scene, which translates to considerable digital transformations. Even when a single additional hue light source was added physically, our model reports that all color transformations – hue, brightness, contrast and saturation – are necessary to match the image of the original scene to the image with the added hue light. Our neural network achieves an average mean squared error loss of 0.5327, suggesting that color transformations alone do not fully capture the changes introduced by an additional hue light source.

\section{Further Details of Experimental Setup}
\label{appx:setup}

To quantify the stability and general behavior of the generated patches, a basic setup was created. The scene to be attacked consisted of a few simple objects: 
\begin{itemize}
    \item a bottle,
    \item a cup,
    \item a small potted plant,
    \item a vase where the potted plant sits in,
    \item a tennis racket,
    \item a spoon,
    \item and a picture of a person.
\end{itemize}

\begin{figure}[h]
    \centering
    \includegraphics[width=0.45\linewidth]{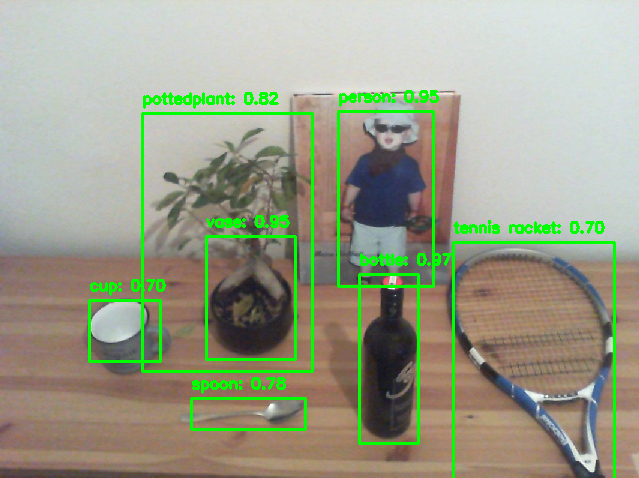}
    \includegraphics[width=0.45\linewidth]{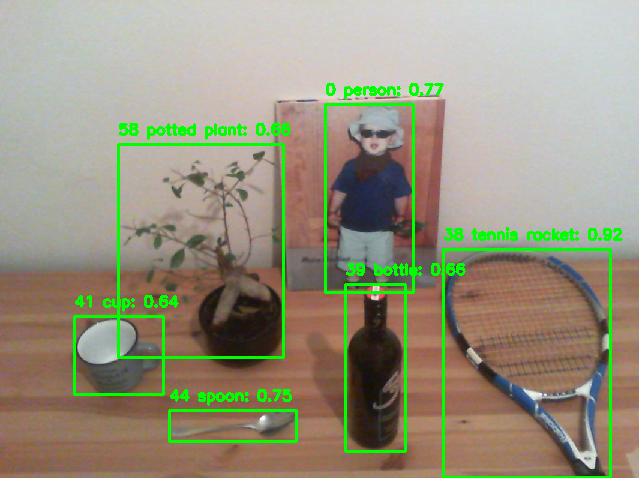}
    \caption{Scene setup with YOLOv3 detections (left) and YOLOv5 detections (right).}
    \label{fig:setup_w_detections}
\end{figure}

A baseline without a patch in the scene and all detected objects using YOLOv3 and YOLOv5 can be seen in \Figref{fig:setup_w_detections}. Occasionally, a dining table is recognized with low confidence by the object detectors. However, this is not included in the ground truth because this object is not actually a dining table, and its recognition depends heavily on the exact position of the camera, which can vary slightly between the experiments. In the case of YOLOv5 detections (\Figref{fig:setup_w_detections} (right)) it is noticeable that the confidence for the vase is very low or it is not detected at all, which is why this object was removed from further observations.

\section{Performance comparison of local and global adversarial patches}
\label{sec:comparison}

This work investigates two different types of attacks: local (using YOLOv5) and global (using YOLOv3). All physical experiments described in Section \ref{sec:exploration}, except for the distance dependence, were repeated for the local patch. Here, we present the results of those experiments and compare the performance of two types of attacks. Important to notice is that evaluation metrics used to measure the efficiency of the attacks in two cases are different. The evaluation metric used for measuring local patch performance is the detection confidence of the attacked object. We interpret this value as follows: the lower the confidence, the better the patch performs. The first two columns in Table~\ref{tab:baselineV5} show the detection confidence values for the standard case, with normal lighting and no patch.

\begin{table}[H]
	\centering
	\begin{tabular}{c||cc|cc}
	\toprule 
		& \multicolumn{2}{c|}{\textbf{without patch}}
		& \multicolumn{2}{c}{\textbf{with patch}} \\ 
	\midrule
		\makecell{\textbf{detection} \\ \textbf{confidence}}            
		& \multicolumn{1}{c|}{\textbf{mean}} 
		& \textbf{std. dev.} 
		& \multicolumn{1}{c|}{\textbf{mean}} 
		& \textbf{std. dev.} \\ 
	\midrule
	\textbf{person} 		& \multicolumn{1}{c|}{0,704}	& 0,029	& \multicolumn{1}{c|}{0,192}	& 0,168	\\ 
	\textbf{cup} 			& \multicolumn{1}{c|}{0,781}	& 0,066	& \multicolumn{1}{c|}{0}		& 0	\\ 
	\textbf{plant}  		& \multicolumn{1}{c|}{0,814}	& 0,028	& \multicolumn{1}{c|}{0,514}	& 0,070 \\ 
	\textbf{bottle}  		& \multicolumn{1}{c|}{0,670}	& 0,072	& \multicolumn{1}{c|}{0}		& 0	\\ 
	\textbf{spoon}  		& \multicolumn{1}{c|}{0,768}	& 0,013	& \multicolumn{1}{c|}{0,162}	& 0,145	\\ 
	\textbf{vase}   		& \multicolumn{1}{c|}{0,141}	& 0,176	& \multicolumn{1}{c|}{0}		& 0	\\ 
	\textbf{tennis racket}  & \multicolumn{1}{c|}{0,915}	& 0,003	& \multicolumn{1}{c|}{0,134}	& 0,162 \\ 
	\bottomrule
	\end{tabular}
	\caption{Confidence of each object with normal lighting and no patch, and normal lighting and patch applied (11cm x 11cm patch size for tennis racket, 7cm x 7cm patch size for all other objects).}
	\label{tab:baselineV5}
\end{table}

\begin{figure}[h]
    \centering
    \includegraphics[width=0.5\textwidth]{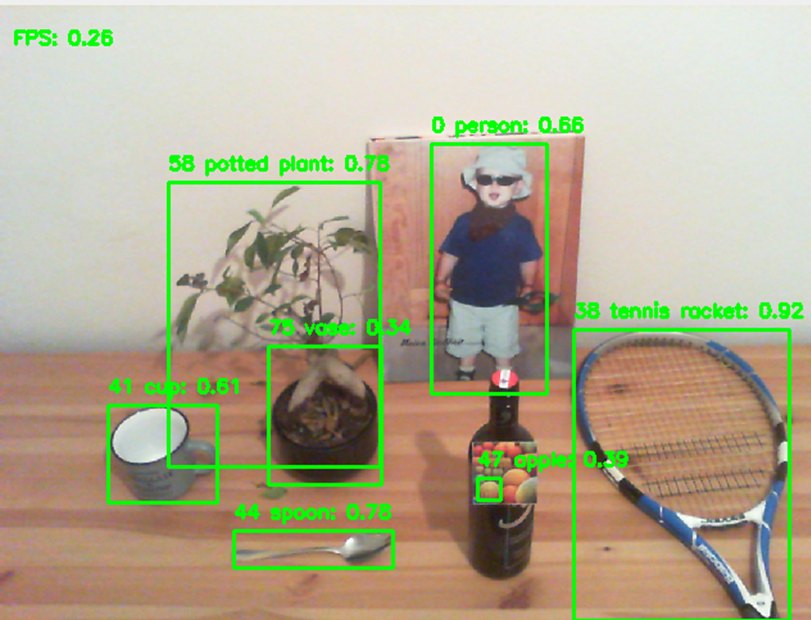}
    \caption{Successful local attack on bottle, 7cm x 7cm patch.}
    \label{fig:ex_local_bottle}
\end{figure}

All objects were attacked locally with a patch approximately 7 cm x 7 cm in size, except for the tennis racket, which required a larger size patch, approximately 11cm x 11cm, to significantly decrease its confidence. The second two columns in Table~\ref{tab:baselineV5} show the baseline confidence for all objects attacked under standard brightness and hue with the patch parallel to the image plane. Each object was attacked individually, with the patch placed overlapping the object. An example is shown in \Figref{fig:ex_local_bottle}.

Due to the nature of these experiments, it was not always possible to conduct them with equal quality for all objects. For instance, applying a local patch with different rotation angles to the potted plant proved to be very difficult. Therefore, only the results for the objects where the patch application could be done most consistently are included here.

\subsection{Rotation dependence}
The results of the rotation experiment with the local patch are quite similar to those with the global patch. The attack performs relatively well up to a rotation angle of 30° around all axes. However, a slight difference is noted: rotation around the z-axis does not affect performance as much as it does for the global patch, indicating that the local patch is a bit more stable in this regard. The recorded confidence values for this experiment are shown in Tables \ref{tab:x_rotation}, \ref{tab:y_rotation}, and \ref{tab:z_rotation}. 

\begin{table}[H]
\centering
\begin{tabular}{c||cc|cc|cc|cc}
\toprule
              & \multicolumn{2}{c|}{\textbf{person}}                    & \multicolumn{2}{c|}{\textbf{cup}}                       & \multicolumn{2}{c|}{\textbf{bottle}}                    & \multicolumn{2}{c}{\textbf{tennis racket}}             \\ 
\midrule
    \makecell{\textbf{rotation around} \\ \textbf{x axis [°]}}            & \multicolumn{1}{c|}{\textbf{mean}} & \textbf{std. dev.} & \multicolumn{1}{c|}{\textbf{mean}} & \textbf{std. dev.} & \multicolumn{1}{c|}{\textbf{mean}} & \textbf{std. dev.} & \multicolumn{1}{c|}{\textbf{mean}} & \textbf{std. dev.} \\ 
\midrule
\textbf{-60} & \multicolumn{1}{c|}{0,64}          & 0,038              & \multicolumn{1}{c|}{0,4}           & 0,157              & \multicolumn{1}{c|}{0,713}         & 0,052              & \multicolumn{1}{c|}{0,686}         & 0,049              \\ 
\textbf{-30} & \multicolumn{1}{c|}{0,374}         & 0,097              & \multicolumn{1}{c|}{0,431}         & 0,16               & \multicolumn{1}{c|}{0,241}         & 0,165              & \multicolumn{1}{c|}{0,6}           & 0,053              \\ 
\textbf{0}   & \multicolumn{1}{c|}{0,192}         & 0,168              & \multicolumn{1}{c|}{0}             & 0                  & \multicolumn{1}{c|}{0}             & 0                  & \multicolumn{1}{c|}{0,134}         & 0,162              \\ 
\textbf{30}  & \multicolumn{1}{c|}{0,188}         & 0,14               & \multicolumn{1}{c|}{0}             & 0                  & \multicolumn{1}{c|}{0}             & 0                  & \multicolumn{1}{c|}{0,766}         & 0,034              \\ 
\textbf{60}  & \multicolumn{1}{c|}{0,374}         & 0,064              & \multicolumn{1}{c|}{0,59}          & 0,085              & \multicolumn{1}{c|}{0,372}         & 0,12               & \multicolumn{1}{c|}{0,87}          & 0,015              \\ 
\bottomrule
\end{tabular}
\caption{Confidence of example objects with standard patch applied over rotation around the x axis.}
\label{tab:x_rotation}
\end{table}

\begin{table}[H]
\centering
\begin{tabular}{c||cc|cc|cc|cc}
\toprule
              & \multicolumn{2}{c|}{\textbf{person}}                    & \multicolumn{2}{c|}{\textbf{cup}}                       & \multicolumn{2}{c|}{\textbf{bottle}}                    & \multicolumn{2}{c}{\textbf{tennis racket}}             \\ 
\midrule
    \makecell{\textbf{rotation around} \\ \textbf{y axis [°]}}            & \multicolumn{1}{c|}{\textbf{mean}} & \multicolumn{1}{c|}{\textbf{std. dev.}} & \multicolumn{1}{c|}{\textbf{mean}} & \multicolumn{1}{c|}{\textbf{std. dev.}} & \multicolumn{1}{c|}{\textbf{mean}} & \multicolumn{1}{c|}{\textbf{std. dev.}} & \multicolumn{1}{c|}{\textbf{mean}} & \multicolumn{1}{c}{\textbf{std. dev.}} \\ 
\midrule
\textbf{-60} & \multicolumn{1}{c|}{0,613}         & 0,034                                   & \multicolumn{1}{c|}{0,659}         & 0,144                                   & \multicolumn{1}{c|}{0,59}          & 0,08                                    & \multicolumn{1}{c|}{0,75}          & 0,114                                   \\ 
\textbf{-30} & \multicolumn{1}{c|}{0,458}         & 0,058                                   & \multicolumn{1}{c|}{0,137}         & 0,178                                   & \multicolumn{1}{c|}{0}             & 0                                       & \multicolumn{1}{c|}{0,163}         & 0,183                                   \\ 
\textbf{0}   & \multicolumn{1}{c|}{0,192}         & 0,168                                   & \multicolumn{1}{c|}{0}             & 0                                       & \multicolumn{1}{c|}{0}             & 0                                       & \multicolumn{1}{c|}{0,134}         & 0,162                                   \\ 
\textbf{30}  & \multicolumn{1}{c|}{0,376}         & 0,088                                   & \multicolumn{1}{c|}{0}             & 0                                       & \multicolumn{1}{c|}{0}             & 0                                       & \multicolumn{1}{c|}{0,334}         & 0,012                                   \\ 
\textbf{60}  & \multicolumn{1}{c|}{0,6}           & 0,03                                    & \multicolumn{1}{c|}{0,14}          & 0,161                                   & \multicolumn{1}{c|}{0,64}          & 0,058                                   & \multicolumn{1}{c|}{0,85}          & 0,018                                   \\ 
\bottomrule
\end{tabular}
\caption{Confidence of example objects with standard patch applied over rotation around the y axis.}
\label{tab:y_rotation}
\end{table}

\begin{table}[H]
\centering
\begin{tabular}{c||cc|cc|cc|cc}
\toprule
            & \multicolumn{2}{c|}{\textbf{person}}                                         & \multicolumn{2}{c|}{\textbf{cup}}                                            & \multicolumn{2}{c|}{\textbf{bottle}}                                         & \multicolumn{2}{c}{\textbf{tennis racket}}                                  \\
\midrule
    \makecell{\textbf{rotation around} \\ \textbf{z axis [°]}}            & \multicolumn{1}{c|}{\textbf{mean}} & \multicolumn{1}{c|}{\textbf{std. dev.}} & \multicolumn{1}{c|}{\textbf{mean}} & \multicolumn{1}{c|}{\textbf{std. dev.}} & \multicolumn{1}{c|}{\textbf{mean}} & \multicolumn{1}{c|}{\textbf{std. dev.}} & \multicolumn{1}{c|}{\textbf{mean}} & \multicolumn{1}{c}{\textbf{std. dev.}} \\ 
\midrule
\textbf{-60} & \multicolumn{1}{c|}{0,596}         & 0,043                                   & \multicolumn{1}{c|}{0,69}          & 0,04                                    & \multicolumn{1}{c|}{0}             & 0                                       & \multicolumn{1}{l|}{0,778}         & 0,032                                   \\ 
\textbf{-30} & \multicolumn{1}{c|}{0,517}         & 0,051                                   & \multicolumn{1}{c|}{0,013}         & 0,062                                   & \multicolumn{1}{c|}{0}             & 0                                       & \multicolumn{1}{l|}{0,199}         & 0,168                                   \\ 
\textbf{0}   & \multicolumn{1}{c|}{0,192}         & 0,168                                   & \multicolumn{1}{c|}{0}             & 0                                       & \multicolumn{1}{c|}{0}             & 0                                       & \multicolumn{1}{l|}{0,134}         & 0,162                                   \\ 
\textbf{30}  & \multicolumn{1}{c|}{0,5}           & 0,053                                   & \multicolumn{1}{c|}{0,044}         & 0,107                                   & \multicolumn{1}{c|}{0}             & 0                                       & \multicolumn{1}{l|}{0,414}         & 0,079                                   \\ 
\textbf{60}  & \multicolumn{1}{c|}{0,488}         & 0,042                                   & \multicolumn{1}{c|}{0,058}         & 0,126                                   & \multicolumn{1}{c|}{0}             & 0                                       & \multicolumn{1}{l|}{0,215}         & 0,166                                   \\ 
\toprule
\end{tabular}
\caption{Confidence of example objects with standard patch applied over rotation around the z axis.}
\label{tab:z_rotation}
\end{table}

\subsection{Size dependence}
Varying the size of the patch leads to the straightforward conclusion that a larger patch significantly improves the performance, as shown in Table \ref{tab:size}. These results align with the experiments performed with the global patch. However, there is one thing different in local attacks setup we should address to. Due to the nature of a local patch, it has to overlap the attacked object. This leads to targeted object becomes less and less visible with an increase in patch size. All recorded confidences (that are not \enquote{N/A}) have been verified to be caused by the patch (and not the inability to see the targeted object) by applying a control patch. This control patch, generated from random noise, has no adversarial properties. If the confidence with the control patch is higher than with the local patch by at least 0.3, it is considered a valid data point. Otherwise, too much of the object is overlapped to be seen and recognized, even without the adversarial properties of the local patch.

\begin{table}[H]
\centering
\begin{tabular}{c||cc|cc|cc|cc}

\toprule
                             & \multicolumn{2}{c|}{\textbf{person}}                                         & \multicolumn{2}{c|}{\textbf{plant}}                                          & \multicolumn{2}{c|}{\textbf{bottle}}                                         & \multicolumn{2}{c}{\textbf{tennis racket}} \\ 
\midrule
\textbf{patch size {[}cm{]}} & \multicolumn{1}{c|}{\textbf{mean}} & \multicolumn{1}{c|}{\textbf{std. dev.}} & \multicolumn{1}{c|}{\textbf{mean}} & \multicolumn{1}{c|}{\textbf{std. dev.}} & \multicolumn{1}{c|}{\textbf{mean}} & \multicolumn{1}{c|}{\textbf{std. dev.}} & \multicolumn{1}{c|}{\textbf{mean}} & \multicolumn{1}{c}{\textbf{std. dev.}} \\ 
\midrule
\textbf{4x4}                 & \multicolumn{1}{c|}{0,68}          & 0,028                                   & \multicolumn{1}{c|}{0,8}           & 0,019                                   & \multicolumn{1}{c|}{0,55}          & 0,1                                     & \multicolumn{1}{c|}{0,874}         & 0,019                                   \\ 
\textbf{7x7}                 & \multicolumn{1}{c|}{0,166}         & 0,16                                    & \multicolumn{1}{c|}{0,52}          & 0,06                                    & \multicolumn{1}{c|}{0,003}         & 0,03                                    & \multicolumn{1}{c|}{0,86}          & 0,018                                   \\ 
\textbf{11x11}               & \multicolumn{1}{c|}{0}             & 0                                       & \multicolumn{1}{c|}{0,457}         & 0,065                                   & \multicolumn{1}{c|}{0}             & 0                                       & \multicolumn{1}{c|}{0,088}         & 0,139                                   \\ 
\textbf{16x16}               & \multicolumn{1}{c|}{0}             & 0                                       & \multicolumn{1}{c|}{N/A}           &                    ---                   & \multicolumn{1}{c|}{N/A}           &                  ---                    & \multicolumn{1}{c|}{0}             & 0                                       \\ 
\bottomrule
\end{tabular}
\caption{Confidence of example objects with different patch sizes.}
\label{tab:size}
\end{table}

\subsection{Brightness dependence}
The variable brightness experiments show the most significant difference in performance between local and global patches. While the effectiveness of global patch decreases with increased brightness, the local patch performs better under brighter light for most objects, as can be seen in Table \ref{tab:brightness}. This improved performance of the local patch at high brightness levels could be due to the lack of light spots, resulting in less clipping from high exposure.  However, the results for tennis racket suggest an inverted behavior than for the other objects, similar to the global patch experiment. This set of experiments would require further testing to investigate different behaviours.

\begin{table}[H]
\begin{tabular}{c||cc|cc|cc|cc}
\toprule
                              & \multicolumn{2}{c|}{\textbf{person}}                    & \multicolumn{2}{c|}{\textbf{plant}}                     & \multicolumn{2}{c|}{\textbf{bottle}}                    & \multicolumn{2}{c}{\textbf{tennis racket}}             \\
\midrule
\textbf{brightness {[}lux{]}} & \multicolumn{1}{c|}{\textbf{mean}} & \textbf{std. dev.} & \multicolumn{1}{c|}{\textbf{mean}} & \textbf{std. dev.} & \multicolumn{1}{c|}{\textbf{mean}} & \textbf{std. dev.} & \multicolumn{1}{c|}{\textbf{mean}} & \textbf{std. dev.} \\ 
\midrule
\textbf{4}                    & \multicolumn{1}{c|}{0,625}         & 0,0056             & \multicolumn{1}{c|}{0,5}           & 0,11               & \multicolumn{1}{c|}{0,367}         & 0,191              & \multicolumn{1}{c|}{N/A}           & 0                  \\ 
\textbf{6}                    & \multicolumn{1}{c|}{0,541}         & 0,106              & \multicolumn{1}{c|}{0,498}         & 0,07               & \multicolumn{1}{c|}{0,258}         & 0,16               & \multicolumn{1}{c|}{0}             & 0                  \\ 
\textbf{10}                   & \multicolumn{1}{c|}{0,404}         & 0,121              & \multicolumn{1}{c|}{0,504}         & 0,83               & \multicolumn{1}{c|}{0,067}         & 0,128              & \multicolumn{1}{c|}{0,06}          & 0,12               \\ 
\textbf{15}                   & \multicolumn{1}{c|}{0,162}         & 0,167              & \multicolumn{1}{c|}{0,523}         & 0,066              & \multicolumn{1}{c|}{0}             & 0                  & \multicolumn{1}{c|}{0,377}         & 0,156              \\ 
\textbf{25}                   & \multicolumn{1}{c|}{0,166}         & 0,169              & \multicolumn{1}{c|}{0,523}         & 0,058              & \multicolumn{1}{c|}{0}             & 0                  & \multicolumn{1}{c|}{0,601}         & 0,069              \\ 
\textbf{35}                   & \multicolumn{1}{c|}{0,043}         & 0,104              & \multicolumn{1}{c|}{0,528}         & 0,059              & \multicolumn{1}{c|}{0}             & 0                  & \multicolumn{1}{c|}{0,7}           & 0,042              \\ 
\textbf{45}                   & \multicolumn{1}{c|}{0,036}         & 0,0065             & \multicolumn{1}{c|}{0,517}         & 0,04               & \multicolumn{1}{c|}{0}             & 0                  & \multicolumn{1}{c|}{0,75}          & 0,023              \\ 
\textbf{50}                   & \multicolumn{1}{c|}{0,02}          & 0,072              & \multicolumn{1}{c|}{0,42}          & 0,047              & \multicolumn{1}{c|}{0}             & 0                  & \multicolumn{1}{c|}{0,75}          & 0,025              \\ 
\textbf{61}                   & \multicolumn{1}{c|}{0,009}         & 0,052              & \multicolumn{1}{c|}{0,378}         & 0,046              & \multicolumn{1}{c|}{0}             & 0                  & \multicolumn{1}{c|}{0,71}          & 0,031              \\ 
\bottomrule
\end{tabular}
\caption{Confidence of example objects with different brightness values.}
\label{tab:brightness}
\end{table}

\subsection{Hue dependence}
\begin{figure}[H]
    \centering
    \includegraphics[width=.5\linewidth]{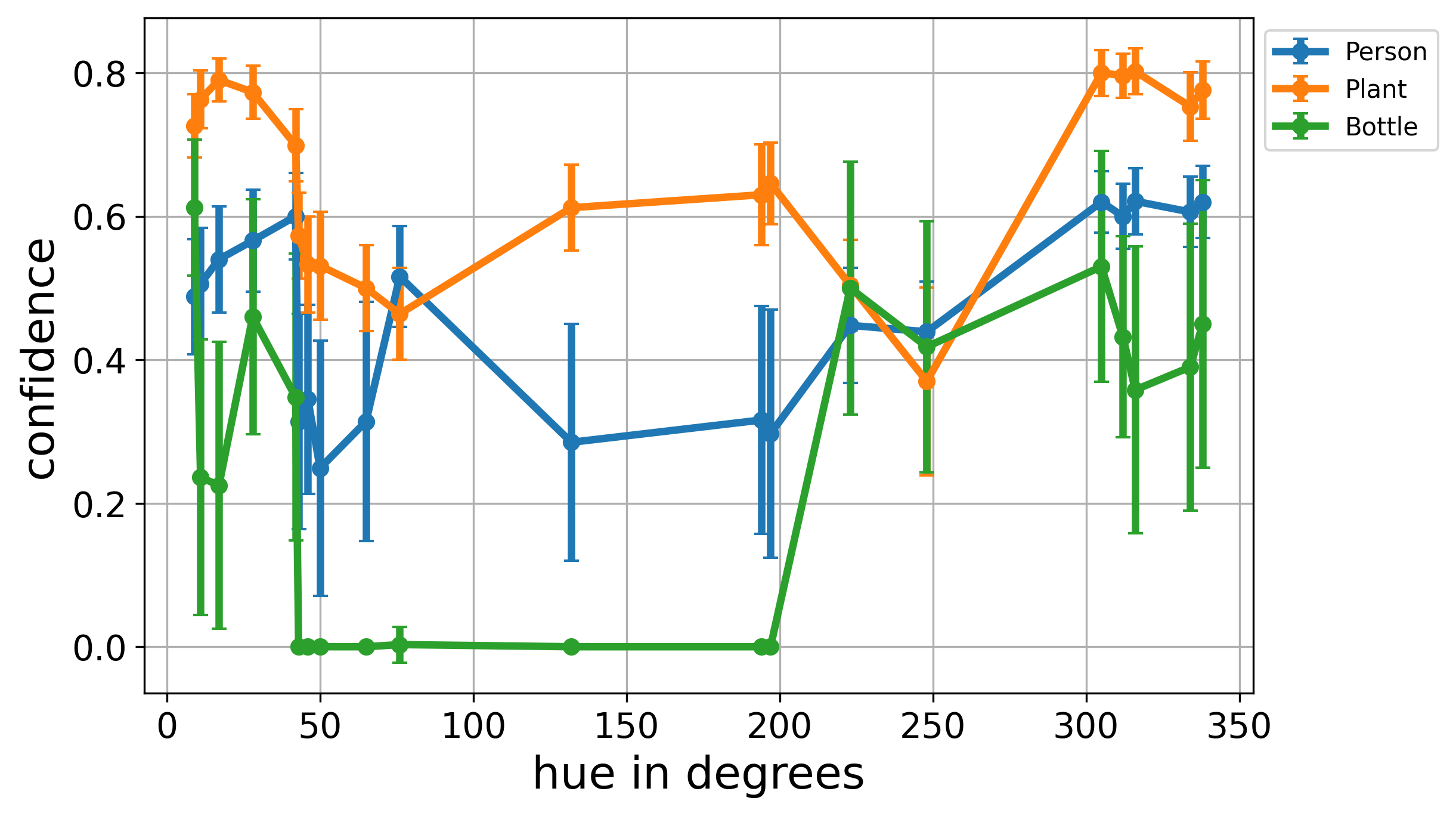}
    \caption{Confidence over hue values of person, plant and bottle as an object under attack by the patch.}
    \label{fig:conf_O_hue_personplantbottle}
\end{figure}

The results of the hue experiment with the local patch are shown in Table~\ref{tab:hue} and \Figref{fig:conf_O_hue_personplantbottle}. The findings are very similar to the global patch experiment, with patch performance decreasing at similar hue values. The \Figref{fig:conf_O_hue_personplantbottle} is less clear that the appropriate one for the global patch, because detection confidence is a less sophisticated metric for comparing patch performance than mean average precision. The plot still shows a clear trend: patch performance worsens at very low hue values (10$^\circ$ to 30$^\circ$) and high hue values above 300$^\circ$, while it performs quite well between 40$^\circ$ and 200$^\circ$/250$^\circ$, depending on the object.

\begin{table}[H]
\centering
\begin{tabular}{c||cc|cc|cc}
\toprule
                     & \multicolumn{2}{c|}{\textbf{person}}                    & \multicolumn{2}{c|}{\textbf{plant}}                     & \multicolumn{2}{c}{\textbf{bottle}}                    \\ 
\midrule
\textbf{hue {[}°{]}} & \multicolumn{1}{c|}{\textbf{mean}} & \textbf{std. dev.} & \multicolumn{1}{c|}{\textbf{mean}} & \textbf{std. dev.} & \multicolumn{1}{c|}{\textbf{mean}} & \textbf{std. dev.} \\ 
\midrule
\textbf{9}           & \multicolumn{1}{c|}{0,488}         & 0,08               & \multicolumn{1}{c|}{0,726}         & 0,044              & \multicolumn{1}{c|}{0,612}         & 0,095              \\ 
\textbf{11}          & \multicolumn{1}{c|}{0,506}         & 0,078              & \multicolumn{1}{c|}{0,763}         & 0,04               & \multicolumn{1}{c|}{0,236}         & 0,192              \\ 
\textbf{17}          & \multicolumn{1}{c|}{0,54}          & 0,074              & \multicolumn{1}{c|}{0,79}          & 0,03               & \multicolumn{1}{c|}{0,225}         & 0,2                \\ 
\textbf{28}          & \multicolumn{1}{c|}{0,566}         & 0,071              & \multicolumn{1}{c|}{0,773}         & 0,037              & \multicolumn{1}{c|}{0,46}          & 0,164              \\ 
\textbf{42}          & \multicolumn{1}{c|}{0,6}           & 0,06               & \multicolumn{1}{c|}{0,699}         & 0,05               & \multicolumn{1}{c|}{0,348}         & 0,2                \\ 
\textbf{43}          & \multicolumn{1}{c|}{0,314}         & 0,15               & \multicolumn{1}{c|}{0,573}         & 0,06               & \multicolumn{1}{c|}{0}             & 0                  \\ 
\textbf{46}          & \multicolumn{1}{c|}{0,345}         & 0,132              & \multicolumn{1}{c|}{0,533}         & 0,067              & \multicolumn{1}{c|}{0}             & 0                  \\ 
\textbf{50}          & \multicolumn{1}{c|}{0,249}         & 0,178              & \multicolumn{1}{c|}{0,531}         & 0,075              & \multicolumn{1}{c|}{0}             & 0                  \\ 
\textbf{65}          & \multicolumn{1}{c|}{0,314}         & 0,167              & \multicolumn{1}{c|}{0,5}           & 0,06               & \multicolumn{1}{c|}{0}             & 0                  \\ 
\textbf{76}          & \multicolumn{1}{c|}{0,516}         & 0,07               & \multicolumn{1}{c|}{0,464}         & 0,064              & \multicolumn{1}{c|}{0,003}         & 0,025              \\ 
\textbf{132}         & \multicolumn{1}{c|}{0,285}         & 0,165              & \multicolumn{1}{c|}{0,612}         & 0,06               & \multicolumn{1}{c|}{0}             & 0                  \\ 
\textbf{194}         & \multicolumn{1}{c|}{0,316}         & 0,159              & \multicolumn{1}{c|}{0,63}          & 0,07               & \multicolumn{1}{c|}{0}             & 0                  \\ 
\textbf{197}         & \multicolumn{1}{c|}{0,297}         & 0,173              & \multicolumn{1}{c|}{0,646}         & 0,057              & \multicolumn{1}{c|}{0}             & 0                  \\ 
\textbf{223}         & \multicolumn{1}{c|}{0,448}         & 0,08               & \multicolumn{1}{c|}{0,504}         & 0,063              & \multicolumn{1}{c|}{0,5}           & 0,176              \\ 
\textbf{248}         & \multicolumn{1}{c|}{0,439}         & 0,07               & \multicolumn{1}{c|}{0,37}          & 0,131              & \multicolumn{1}{c|}{0,418}         & 0,175              \\ 
\textbf{305}         & \multicolumn{1}{c|}{0,62}          & 0,043              & \multicolumn{1}{c|}{0,8}           & 0,032              & \multicolumn{1}{c|}{0,53}          & 0,161              \\ 
\textbf{312}         & \multicolumn{1}{c|}{0,6}           & 0,045              & \multicolumn{1}{c|}{0,796}         & 0,031              & \multicolumn{1}{c|}{0,462}         & 0,14               \\ 
\textbf{316}         & \multicolumn{1}{c|}{0,621}         & 0,046              & \multicolumn{1}{c|}{0,802}         & 0,032              & \multicolumn{1}{c|}{0,358}         & 0,2                \\ 
\textbf{334}         & \multicolumn{1}{c|}{0,606}         & 0,049              & \multicolumn{1}{c|}{0,753}         & 0,048              & \multicolumn{1}{c|}{0,39}          & 0,2                \\ 
\textbf{338}         & \multicolumn{1}{c|}{0,62}          & 0,05               & \multicolumn{1}{c|}{0,776}         & 0,04               & \multicolumn{1}{c|}{0,45}          & 0,2                \\ 
\bottomrule
\end{tabular}
\caption{Confidence of example objects with different hue values.}
\label{tab:hue}
\end{table}

\section{Discovering vulnerabilities of
adversarial patches - Brightness dependence}
\begin{figure}[H]
    \centering
    \includegraphics[width=0.5\linewidth]{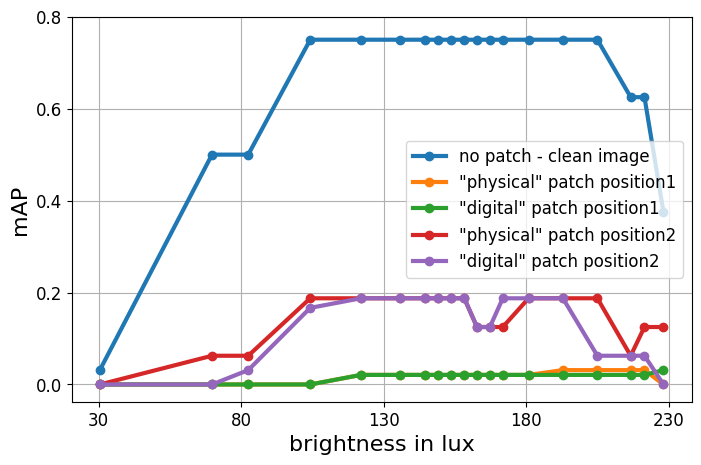}
    \caption{Mean average precision over brightness of the scene for different patch positions in a digital experiment.}
    \label{fig:mAP_img_brightness_multiple_digital}
\end{figure}

\Figref{fig:mAP_img_brightness_multiple_digital} complements the experiments presented in \Secref{brightness_dependence}. The plot shows mean average precision over brightness of the scene for different patch positions (position1 and position2) in digital experiments. These results further confirm the discussion in \Secref{distance_dependence}, that the patch performance is highly dependent on the patch position in the scene. Our experiments indicate that the digital and straightforward imitation of physical patches yield similar performance when placed in identical positions, contrary to our real-world observations. This finding reinforces our assertion that accurately replicating real-world conditions in a digital environment is challenging.

\section{Global adversarial patch color analysis}
\begin{figure}[H]
    \centering
    \includegraphics[width=.5\linewidth]{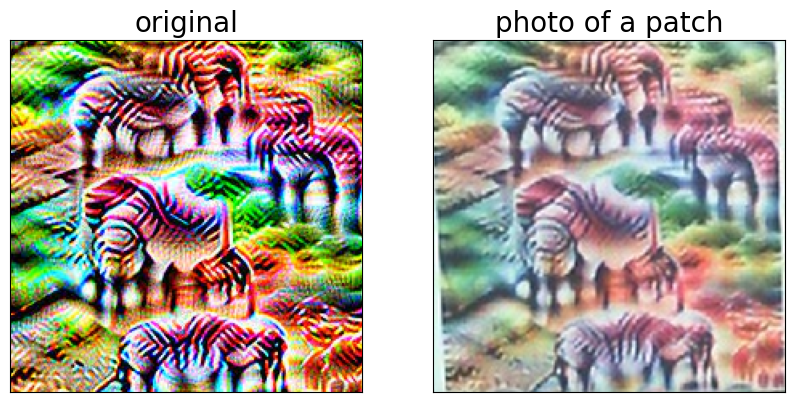}
    \caption{Original patch (left) and photo of the patch captured with our camera (right).}
    \label{fig:original_and_photo_patch]}
\end{figure}

All digital experiments with the global patch are performed with the originally generated version of a patch shown in \Figref{fig:original_and_photo_patch]} (left). The physical experiments use physical patches printed on standard paper. The capture of the physical patch taken with our camera is shown in \Figref{fig:original_and_photo_patch]} (right). As can be seen, the "quality" of the physical patch is highly dependant on environmental conditions and the camera being used. Also, the picture of the patch on \Figref{fig:original_and_photo_patch]} (right) is taken under normal daylight and without any filters or additional obstructions, one can noticed that the colors are less intense and that the edges are less sharp.

\begin{figure}[h]
    \centering
    \includegraphics[width=.4\linewidth]{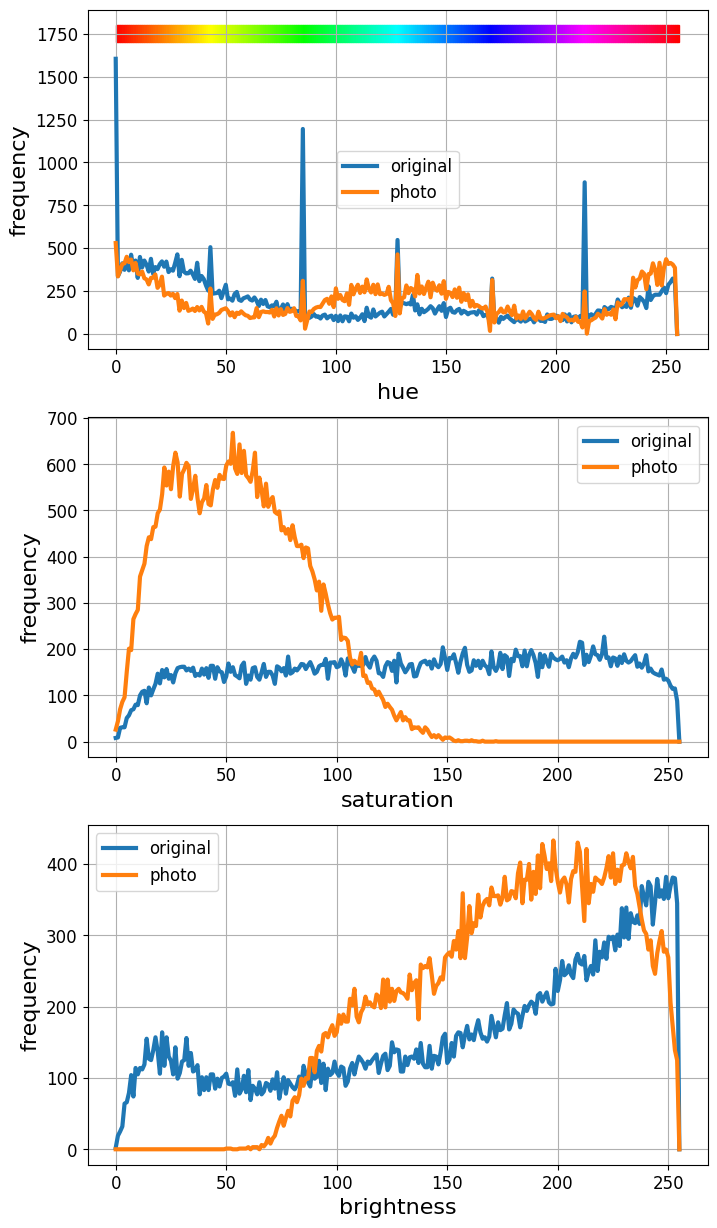}
    \includegraphics[width=.4\linewidth]{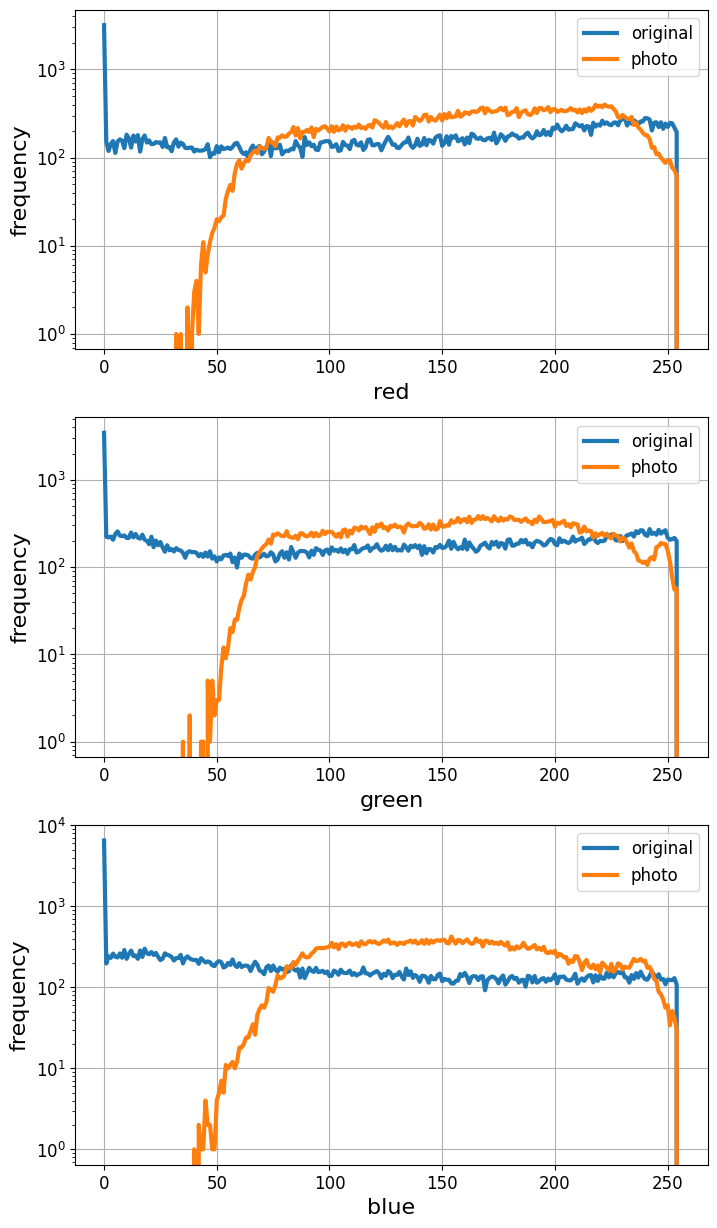}
    \caption{HSV patch analysis (left) and RGB patch analysis (right).}
    \label{fig:original_and_photo_patch_analysis]}
\end{figure}

We performed patch color analysis in HSV (stands for Hue, Saturation and Value -- brightness) color space, and in RGB (stands for Red, Green and Blue) color space. Figure \ref{fig:original_and_photo_patch_analysis]} shows the distribution of different values of hue, saturation, and brightness for both the original patch and photo of the patch (left), and the distribution of different values of red, green, and blue colors for both the original patch and a photo of the patch (right). Our results confirm the intuitive conclusion stated above, that colors in a photo of a patch are less saturated, and support our claim that the transformation from the physical to the digital world is not easy. 

\end{document}